\documentclass[10pt, journal]{IEEEtran}
\usepackage[utf8]{inputenc}
\usepackage[exponent-product=\cdot,detect-all,detect-inline-family=text]{siunitx}
\usepackage[breaklinks]{hyperref}
\hypersetup{hidelinks}
\usepackage{ 
	amsfonts, 
	amssymb, 
	mathtools,
	amsthm,
	siunitx}
\usepackage{graphicx}

\usepackage{pgfplots, pgfplotstable, eso-pic}
\usepackage{multirow}
\usepackage{booktabs}

\usepackage[english,
	ruled,
	]{algorithm2e}
	
\usepackage[nospace]{cite}

\newcommand{\bfa}{\mathbf{a}}

\newcommand{\bbfa}{\bar{\bfa}}

\newcommand{\hbfa}{\hat{\bfa}}
\newcommand{\bfA}{\mathbf{A}}
\newcommand{\bfb}{\mathbf{b}}
\newcommand{\hbfb}{\hat{\mathbf{b}}}

\newcommand{\bfB}{\mathbf{B}}

\newcommand{\bfe}{\mathbf{e}}

\newcommand{\bfF}{\mathbf{F}}

\newcommand{\bfG}{\mathbf{G}}

\newcommand{\bfg}{\mathbf{g}}

\newcommand{\bfI}{\mathbf{I}}

\newcommand{\calJ}{\mathbf{\mathcal{J}}}

\newcommand{\calN}{\mathcal{N}}

\newcommand{\bfp}{\mathbf{p}}
\newcommand{\hbfp}{\hat{\mathbf{p}}}
\newcommand{\dbfp}{\dot{\bfp}}

\newcommand{\bfQ}{\mathbf{Q}}

\newcommand{\calQ}{\mathcal{Q}}

\newcommand{\bfR}{\mathbf{R}}

\newcommand{\dbfR}{\dot{\bfR}}
\newcommand{\hbfR}{\hat{\bfR}}

\newcommand{\bbR}{\mathbb{R}}

\newcommand{\bfS}{\mathbf{S}}

\newcommand{\bfT}{\mathbf{T}}
\newcommand{\hbfT}{\mathbf{\hat{T}}}

\newcommand{\bfv}{\mathbf{v}}

\newcommand{\hbfv}{\hat{\bfv}}

\newcommand{\dbfv}{\dot{\bfv}}

\newcommand{\boeta}{\boldsymbol{\eta}}
\newcommand{\boGamma}{\boldsymbol{\Gamma}}
\newcommand{\dboGamma}{\dot{\boldsymbol{\Gamma}}}
\newcommand{\boomega}{\boldsymbol{\omega}}

\newcommand{\hboomega}{\hat{\boomega}}
\newcommand{\boOmega}{\boldsymbol{\Omega}}

\newcommand{\bonu}{\boldsymbol{\nu}}
\newcommand{\bophi}{\boldsymbol{\phi}}

\newcommand{\boPhi}{\boldsymbol{\Phi}}
\newcommand{\borho}{\boldsymbol{\rho}}

\newcommand{\boSigma}{\boldsymbol{\Sigma}}

\newcommand{\boUpsilon}{\boldsymbol{\Upsilon}}
\newcommand{\hboUpsilon}{\hat{\boldsymbol{\Upsilon}}}
\newcommand{\boxi}{\boldsymbol{\xi}}

\newcommand{\bfzero}{\mathbf{0}}

\newcommand{\mc}{\mathrm{mc}}
\newcommand{\equal}[1]{\ensuremath{\stackrel{\eqref{#1}}{=}}}

\newcommand{\simeqbis}[1]{\ensuremath{\stackrel{\text{(#1)}}{\simeq}}}

\DeclareMathOperator{\cov}{cov}

\DeclareMathOperator{\diag}{diag}

\DeclareMathOperator{\Ad}{Ad}

\DeclareMathOperator{\trace}{trace}
\DeclareMathOperator{\E}{E}

\newtheoremstyle{mystyle}
{}
{}
{\itshape}
{}
{\bfseries}
{.}
{ }
{}
\newtheorem{rmk}{Remark}

\newcommand{\enbleu}[1]{{\color{black}{#1}}}

\title{Associating Uncertainty to Extended Poses for on Lie Group IMU  Preintegration with Rotating Earth}

\author{Martin Brossard, Axel Barrau, Paul Chauchat, and Silvère Bonnabel
\thanks{M. Brossard, A. Barrau and S. Bonnabel are with MINES ParisTech, PSL Research University, Centre for Robotics, 60 Boulevard Saint-Michel, 75006 Paris, France. A. Barrau is with Safran Tech, Groupe Safran, Rue des Jeunes Bois-Ch\^ateaufort, 78772, Magny Les Hameaux Cedex, France. P. Chauchat is with ISAE, 31400 Toulouse, France. S. Bonnabel is with University of New Caledonia, ISEA, 98851, Noumea Cedex, New Caledonia. (e-mail: \href{mailto:martin.brossard@mines-paristech.fr}{martin.brossard@mines-paristech.fr}; \href{mailto:axel.barrau@safrangroup.com}{axel.barrau@safrangroup.com};
\href{mailto:paul.chauchat@isae-supaero.fr}{paul.chauchat@isae-supaero.fr}; \href{mailto:silvere.bonnabel@unc.nc}{silvere.bonnabel@unc.nc})}}
\begin{document}
\bstctlcite{IEEEexample:BSTcontrol}
\maketitle
\begin{abstract}
The recently introduced matrix group $\mathbf{SE_2(3)}$ provides a  5$\mathbf{\times}$5 matrix representation  for the  orientation, velocity and position of an object  in the 3-D space, a triplet we call ``extended pose''. In this paper we build on this group to develop a theory to associate uncertainty with extended poses represented by 5$\mathbf{\times}$5 matrices. Our approach is particularly suited to describe how uncertainty propagates when the extended pose represents the state of an Inertial Measurement Unit (IMU). In particular it allows  revisiting the theory of IMU preintegration on manifold and  reaching a further theoretic level in this field. Exact preintegration formulas that account for rotating Earth, that is, centrifugal force and Coriolis force, are derived as a byproduct, and the factors are shown to be more accurate. The approach is validated through extensive simulations and applied to sensor-fusion where a loosely-coupled fixed-lag smoother fuses IMU and LiDAR on   one hour long   experiments using our experimental car. It shows how handling rotating Earth may be beneficial for long-term navigation within incremental smoothing algorithms.
\end{abstract}
\section{Introduction}\label{sec:intro}
The main contribution of this paper is to provide practical techniques to associate uncertainty to the triplet 
\begin{align*}
    \left\{\text{orientation $\bfR \in SO(3)$, velocity $\bfv \in \bbR^3$, position $\bfp \in \bbR^3$}\right\}
\end{align*}
of a moving platform equipped with an Inertial Measurement Unit (IMU) for use in navigation and estimation problems.

A 4$\times$4 transformation matrix may model the pose of a platform, i.e. its orientation and its position. This has been popularized in computer vision \cite{hartley_multiple_2004}, probabilistic robotics \cite{barfoot_state_2017} and pose-graph SLAM \cite{dellaert_factor_2017}. In particular, Gaussian distributions in exponential coordinates of the Lie group $SE(3)$ provide practical tools for propagating pose uncertainty or fusing multiple measurements of a pose in one estimate, \enbleu{see the pioneering work of  \cite{yunfeng_wang_nonparametric_2008} and its follow-up \cite{barfoot_associating_2014}.}

\begin{figure}
    \centering
    \includegraphics{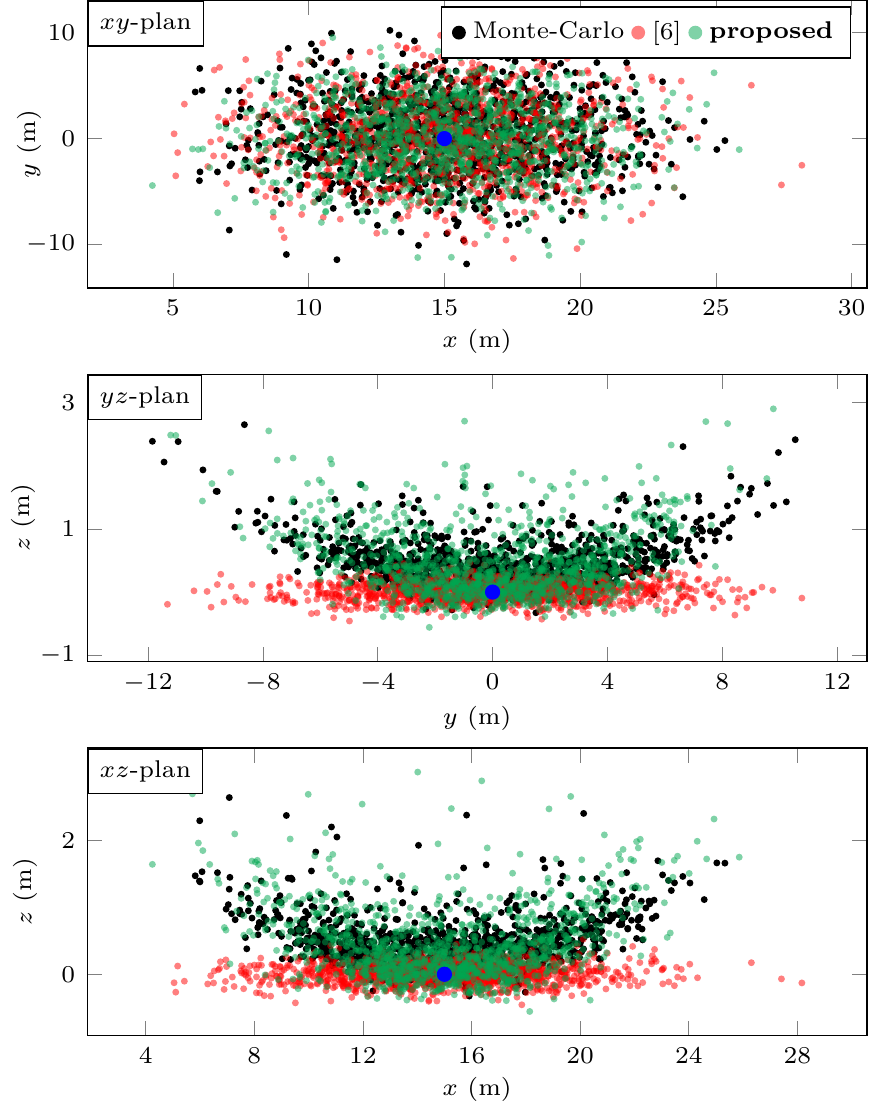}
    \caption{
    Comparison between uncertainty propagation models of an extended pose. An IMU moves from origin to the right at constant translational speed during \SI{10}{s} and stops at the blue dot (15, 0, 0). We preintegrate noisy IMU measurements in a factor and generate a dispersion of the factor belief at the end-point. The true dispersion (black dots) is based on Monte-Carlo using the actual IMU noise model, green dots are generated through the proposed uncertainty model and red dots are generated using state-of-the-art \cite{forster_-manifold_2017}. The true distribution is curved in $xz$- and $yz$-planes, and only the proposed scheme captures this effect and agrees with true dispersion. \label{fig:intro}} 
\end{figure}

When one uses an IMU, a sensor that embeds gyrometers and accelerometers, one manipulates the pose of the sensor and its velocity that we refer in this paper to as an \emph{extended pose}. An extended pose cannot be modeled as an element of $SE(3)$. As a result, the IMU propagation equations are not amenable to the framework of \cite{barfoot_associating_2014,yunfeng_wang_nonparametric_2008}. 

Besides, the theory of IMU preintegration \cite{forster_-manifold_2017,forster_imu_2015} allows defining a  unique factor between two keyframes from a sequence of inertial measurements, independently from the current estimate. In this field  ``\emph{it is of paramount importance to accurately model the noise covariance \emph{[of the preintegrated factor]}}" \cite{forster_-manifold_2017}, i.e. the factor belief. Figure \ref{fig:intro} illustrates how the uncertainty representation of an extended pose advocated in this paper correctly computes the underlying factor belief. 

In this work, which is an extension of   preliminary   results \cite{barrau_mathematical_2020}, we define 5$\times$5 matrices to model  extended poses,  and show how   our approach allows transposing the framework dealing with the Lie group approach to pose uncertainty to the context of IMUs. Our contributions are as follows: 
\begin{enumerate}
    \item we show how to  propagate uncertainty of an extended pose through the IMU kinematic model, where we provide a substantial extension of the results of \cite{barfoot_associating_2014} to position, orientation plus velocity when using IMUs;
    \item we address IMU preintegration. This paper provides a theoretical framework for preintegrating IMU on the Lie group of extended poses. It notably simplifies and improves the preintegration on manifold \cite{forster_-manifold_2017}, provides a rigorous treatment of Coriolis force, and handles IMU biases slightly more accurately than the approach of \cite{forster_-manifold_2017};
    \item we numerically demonstrate the efficiency of our theoretical framework through massive simulations and real experiments;
    \item we implement a loosely-coupled approach based on iSAM2 \cite{kaess_isam2_2012} that obtains reliable estimates for one hour long sequences acquired with our car equipped with an IMU, and a LiDAR that outputs relative translations;
    \item an accompanying set of Python scripts to implement many of the key equations and regenerate the plots in the paper, and our GTSAM fork, are downloadable from \texttt{\url{https://github.com/mbrossar/SE2-3-}};
    \item \enbleu{the supplementary material\footnote{For a public link, please see \texttt{\url{https://github.com/mbrossar/SE2-3-/raw/master/supplementary_material.pdf}}.} that accompanies the present paper contains detailed proofs along with comprehensive technical derivations.}
\end{enumerate}

\subsection{Relations with $SE(3)$ based Uncertainty Propagation}
One contribution of \cite{yunfeng_wang_nonparametric_2008}, and \cite{barfoot_associating_2014} also, consists in compounding pose uncertainties based on a discrete-time integration of the $SE(3)$ kinematic equation as
\begin{align}
\bfT_{i+1} = \bfT_i \boUpsilon_i, \label{eq:com_pos}
\end{align}
where $\bfT_i$ is a global pose and $\boUpsilon_i$ is a local increment, a scheme which is not applicable to IMU kinematic equations. We generalize \eqref{eq:com_pos} and propagate an \emph{extended pose} as
\begin{align}
    \bfT_{i+1} = \boGamma_i \Phi(\bfT_i)\boUpsilon_i,\label{eq:com_ext_pos}
\end{align}
where $\bfT_i$ is a global extended pose, $\boGamma_i$ is a global increment, $\boUpsilon_i$ a local increment, and $\Phi(\cdot)$ a function that depends \emph{only} on the time between instants $i$ and $i+1$. The scheme \eqref{eq:com_pos}  models a robot driving in the environment where $\boUpsilon_i$ is given by, e.g., differential wheel speeds or visual relative pose estimates. To derive kinematic  model \eqref{eq:com_ext_pos}, we build on a novel representation of extended poses by 5$\times$5 matrices, namely the group  $SE_2(3)$, introduced in the recent paper \cite{barrau_invariant_2017} for    navigation.

To express how pose \emph{uncertainty} evolves, \cite{barfoot_associating_2014} associates uncertainty to poses as
\begin{align}
    \bfT_i= \hbfT_i \exp(\boxi_i), \quad \boUpsilon_i=\hboUpsilon_i \exp(\boeta_i), \label{eq:uncer}
\end{align}
where $\hbfT_i$ and $\hboUpsilon_i$ are noise-free variables, $\exp(\cdot)$ is the $SE(3)$ exponential map, and $\boxi_i\sim \calN(\bfzero_6,\boSigma_i)$ and $\boeta_i\sim \calN(\bfzero_6,\bfQ)$ are uncertainties, a.k.a. errors or noises. In a probabilistic context, the compound \eqref{eq:com_pos} is then rewritten as 
\begin{align}
    \hbfT_{i+1}\exp(\boxi_{i+1}) = \hbfT_i \exp(\boxi_i) \hboUpsilon_i \exp(\boeta_i). \label{eq:com_pos2}
\end{align}
Given nominal values and associated uncertainties $\{\hbfT_i,\boSigma_i\}$, $\{\hboUpsilon_i,\bfQ\}$, \cite{barfoot_associating_2014} shows how to compute $\{\hbfT_{i+1},\boSigma_{i+1}\}$.

By similarly leveraging \eqref{eq:uncer}, we investigate in this paper how extended pose uncertainties propagate through \eqref{eq:com_ext_pos} as
\begin{align*}
    \hbfT_{i+1}\exp(\boxi_{i+1}) = \boGamma_i \Phi\left(\hbfT_i \exp(\boxi_i)\right) \hboUpsilon_i \exp(\boeta_i),
\end{align*}
where naturally $\exp(\cdot)$ denotes the exponential map of   $SE_2(3)$, which is an extension of $SE(3)$ suited to robot state estimation involving IMUs.  The obtained formulas have very concrete implications for  preintegration, see Section \ref{sec:preintegration}.

\subsection{Links and Differences with Existing Literature}

In robotics, it is well established that estimating uncertain spatial relationships is fundamentally important for state-estimation \cite{smith_representation_1986, park_lie_1995}, robot control \cite{calinon_gaussians_2020}, or active SLAM \cite{rodriguez-arevalo_importance_2018}.

The pioneering works \cite{thrun_real-time_2000, long_banana_2012} notices that mobile robots dispersion under the effect of sensor noise resembles more a ``banana'' than a standard ellipse, which is accurately approximated with Gaussian distributions in Lie exponential coordinates of $SE(2)$ \cite{chirikjian_stochastic_2009-1}. \enbleu{ This paves the ways for defining uncertainty on manifolds, see e.g. \cite{wang_error_2006,wolfe_bayesian_2011}, and for Gaussians on $SE(3)$ \cite{long_banana_2012, wang_error_2006}. \cite{yunfeng_wang_nonparametric_2008} obtained the first results about uncertainty propagation on $SE(3)$ in a general and non-parametric context, which are complemented by the results of \cite{barfoot_associating_2014} in discrete time. The approach  is extended to continuous time  in \cite{wong_data-driven_2020,tang_white-noise--jerk_2019,anderson_full_2015}, and to  correlated pose uncertainties in \cite{mangelson_characterizing_2019}.}

Preintegrating IMU is an alternative to the standard inertial measurement integration which has de facto been adopted in optimization-based estimation framework such as GTSAM \cite{dellaert_factor_2012} and OKVIS \cite{leutenegger_keyframe_2015}. It was initiated by \cite{lupton_visual-inertial-aided_2012} and later improved in \cite{indelman_information_2013,forster_-manifold_2017,forster_imu_2015,carlone_eliminating_2014} notably for avoiding singularities due to the use of Euler  angles. IMU preintegration is adapted, e.g., for legged robot odometry \cite{wisth_preintegrated_2020,hartley_hybrid_2018}, differential drive motion model \cite{deray_joint_2019}, unknown time offset \cite{yang_analytic_2020}, wheel odometry\cite{quan_tightly_2019,zheng_visual_2019}, and covariance preintegration \cite{allak_covariance_2019}. 
This paper generalizes and goes beyond the manifold representation of \cite{forster_-manifold_2017}. Indeed the latter is concerned with the manifold structure of Lie group $SO(3)$ only, and treats the remainder of the state linearly,  whereas we embed the whole state into the group $SE_2(3)$. Beyond the ``on-manifold'' approach  we hence also benefit from the fact the very structure of $SE_2(3)$ proves much more accurate for describing IMU-related equations and uncertainty propagation.  \enbleu{The choice of the $SE_2(3)$ Lie group formalism takes its roots in invariant Kalman filtering, see \cite{barrau_invariant_2017,barrau_invariant_2018}, which has led to more accurate and consistent extended Kalman filters and has yet prompted a high-end industrial product\cite{barrau_alignment_2016}.}

The work of \cite{shen_tightly_2015}, later extended in a visual-inertial navigation system in \cite{qui_vins-mono-2018}, introduces preintegration in its continuous form, \cite{le_gentil_gaussian_2020} proposes asynchronous preintegration with Gaussian processes, \cite{fourie_multi_2017} addresses continuous
preintegration as a higher-order Taylor expansion, \cite{henawy_accurate_2020} describes a scheme based on switched linear systems, and \cite{eckenhoff_direct_2017, eckenhoff_closed-form_2019} provide closed-form expressions for computing analytically the preintegration factors. This work improves the Euler integration scheme of \cite{forster_-manifold_2017} to limit discretization errors, whose integration schemes remain compatible with our approach based on exact discretization.

This paper is an extension of our preliminary conference paper \cite{barrau_mathematical_2020}, see also early ideas in \cite{barrau_linear_2019}, that contains  in-depth discussions, comprehensive technical derivations, numerical and real experiments, and publicly available implementation.

\subsection{Organization of the Paper}
The remainder of the paper is organized as follows. Section \ref{sec:maths} presents the mathematical tools that we use throughout the paper. Section \ref{sec:IMU} recaps our previous results. Section \ref{sec:propagating} shows how to propagate an extended pose and its uncertainty. Section \ref{sec:preintegration} addresses IMU preintegration on flat Earth and Section \ref{sec:coriolis} its extension on rotating Earth. Section \ref{sec:experiments} contains real experiments demonstrating the relevance of the approach. 

\section{Mathematical Preliminaries}\label{sec:maths}
To familiarize with Lie group theory for robotics, please refer to \cite{barfoot_state_2017}, or \cite{chirikjian_stochastic_2009-1, sola_micro_2018}.

\subsection{$SE_2(3)$, the Lie Group of Extended Poses}

Estimating the orientation $\bfR$, velocity $\bfv$ and position $\bfp$ of a rigid body in space is a common problem in robotic navigation and perception. We represent in this paper rotation $\bfR$ with the special orthogonal group 
\begin{align}
    SO(3):=\left\{\bfR \in \bbR^{3\times3} ~|~ \bfR\bfR^T = \bfI_3,~ \det \bfR = 1\right\}.\nonumber
\end{align}

The set of poses may be defined using homogeneous matrices of the special Euclidean group
\begin{align}
SE(3):=\left\{\bfT= \begin{bmatrix}\begin{array}{c|c}\bfR &\bfp\\\midrule\bfzero_3^T &1\end{array}\end{bmatrix} \in \bbR^{4\times4}~\bigg|~\begin{matrix}\bfR \in SO(3)\\ \bfp \in \bbR^3\end{matrix}\right\}.\nonumber
\end{align}
This paper describes extended poses through the following group first introduced to the best of our knowledge in \cite{barrau_invariant_2017}
\begin{align}
SE_2(3):=\left\{\bfT = \begin{bmatrix}\begin{array}{c|cc}\bfR &\bfv&\bfp\\\midrule\bfzero_{2\times3} & \multicolumn{2}{c}{\bfI_2} \end{array} \end{bmatrix} \in \bbR^{5\times5}\Bigg| \begin{matrix}~\bfR \in SO(3)\\\bfv, \bfp  \in \bbR^3\end{matrix}\right\}.\nonumber
\end{align}

$SO(3)$, $SE(3)$ and $SE_2(3)$ are matrix Lie groups where matrix multiplication  provides group composition of two elements, and matrix inverse provides element inverse. The linear operator $\wedge$ maps elements $\boxi\in\bbR^9$ to the Lie algebra of $SE_2(3)$ 
\begin{align}
    \boxi^\wedge &:= \begin{bmatrix}
        \bophi \\ \bonu \\ \borho
    \end{bmatrix}^\wedge 
    := \begin{bmatrix}\begin{array}{c|cc}
        \bophi_\times & \bonu & \borho\\\midrule
        \bfzero_{2\times3} & \multicolumn{2}{c}{\bfzero_{2\times2}}
    \end{array}\end{bmatrix} \in \frak{se}_2(3),
\end{align}
where $\bophi \in \bbR^3$, $\bonu \in \bbR^3$, $\borho \in \bbR^3$, and  
\begin{align}
    \bophi_\times := \begin{bmatrix}\phi_1\\\phi_2\\\phi_3\end{bmatrix}_\times := \begin{bmatrix}
        0 &-\phi_3&\phi_2\\
        \phi_3&0&-\phi_1\\
        -\phi_2&\phi_1&0
    \end{bmatrix}
\end{align}
denotes the skew symmetric matrix associated with cross product with $\bophi$, which maps also a  3$\times$1 vector to an element of $\frak{so}(3)$, the Lie algebra of $SO(3)$.

\subsubsection{Exponential, Logarithm, \& Adjoint Operator}
the exponential map conveniently maps an element $\boxi \in \bbR^9$ to $SE_2(3)$ 
\begin{align}
    \exp(\boxi):= \exp_m(\boxi^\wedge) = \sum_{k=0}^{\infty}\frac{1}{k!}\left(\boxi^\wedge\right)^k,
\end{align}
\enbleu{and its local inverse, the logarithm map, defined as}
\begin{align}
\color{black}\log(\bfT)&\color{black}= \left(\sum_{k=1}^{\infty}(-1)^{k+1} \frac{(\bfT-\bfI_5)^k}{k}\right)^{\vee},
\end{align}
\enbleu{where $\vee$ is the inverse of $\wedge$, enables to map the small perturbation $\exp(\boxi) \in SE_2(3)$ to $\bbR^9$ such that}
\begin{align}
    \boxi = \log\big(\exp\left(\boxi\right)\big).
\end{align}

We conveniently define the adjoint operator
\begin{align}
    \Ad_\bfT := \begin{bmatrix}
        \bfR & \bfzero_{3\times3} & \bfzero_{3\times3} \\
        \bfv_\times \bfR& \bfR & \bfzero_{3\times3} \\
        \bfp_\times \bfR&  \bfzero_{3\times3} & \bfR
    \end{bmatrix}
\end{align}
as an operator acting directly on $\bbR^9$. The following relations prove extremely useful:
\begin{align}
    \bfT \exp(\boxi) \bfT^{-1}& = \exp(\Ad_\bfT \boxi),\label{eq:adjoint}\\
    \Rightarrow\bfT \exp(\boxi) &= \exp(\Ad_\bfT \boxi) \bfT,\label{eq:adjoint2} \\
    \Ad_{\boGamma}\Ad_{\bfT} &= \Ad_{\boGamma \bfT}.\label{eq:adjoint3}
\end{align}

\subsubsection{Baker-Campbell-Hausdorff (BCH) Formula} the BCH formula provides powerful tools to manipulate uncertainty on Lie groups, which can be used to compound two matrix exponentials
\begin{align*}
    \log_m\big(\exp_m(\bfA)\exp_m(\bfB)\big) = \bfA + \bfB + \frac{1}{2}\left[\bfA,\bfB\right]  \nonumber \\
    +\frac{1}{12}\left(\left[\bfA, \left[\bfA,\bfB\right]\right] + \left[\bfB, \left[\bfB,\bfA\right]\right]\right)
    - \frac{1}{24}\left[\bfB, \left[\bfA,\left[\bfA,\bfB\right]\right]\right] + \ldots
\end{align*}
in terms of infinite series, where the Lie bracket is given by
\begin{align*}
    [\bfA,\bfB] := \bfA\bfB - \bfB\bfA.
\end{align*}
In the particular case of $SE_2(3)$, we have
\begin{align}
    [\boxi^\wedge,\boeta^\wedge] := \boxi^\wedge \boeta^\wedge - \boxi^\wedge \boeta^\wedge = \left(\boxi^\curlywedge \boeta\right)^\wedge,
\end{align}
where the ``curlywedge'' $\curlywedge$ operator is defined as 
\begin{align}
    \boxi^\curlywedge := \begin{bmatrix}
        \bophi \\ \bonu \\ \borho
    \end{bmatrix}^\curlywedge 
    := \begin{bmatrix}
        \bophi_\times &\bfzero_{3\times3} &\bfzero_{3\times3} \\
        \bonu_\times & \bophi_\times & \bfzero_{3\times3} \\
        \borho_\times & \bfzero_{3\times3} &\bophi_\times
    \end{bmatrix},
\end{align}
which is 9$\times$9. We note that
\begin{align}
    \log\big(\exp\left(\boxi\right)\exp\left(\boeta\right)\big) = \boxi + \boeta + \frac{1}{2}\boxi^\curlywedge\boeta \nonumber   \\ +\frac{1}{12}\left(\boxi^\curlywedge\boxi^\curlywedge\boeta + \boeta^\curlywedge\boeta^\curlywedge\boxi\right)  -\frac{1}{24}\boeta^\curlywedge\boxi^\curlywedge\boxi^\curlywedge\boeta + \ldots, \label{eq:bchapp}
\end{align}
showing $\exp(\boxi)\exp(\boeta)\neq\exp(\boxi+\boeta)$. We thus require approximations to compute the logarithm of a product of exponentials. If we assume, e.g., that $\boeta$ is a small noise while $\boxi$ is non-negligible, we get
\begin{align}
    \log\big(\exp\left(\boxi\right)\exp\left(\boeta\right)\big) = \boxi + \calJ^{-1}_{\boxi} \boeta + O(\|\boeta\|^2),\label{eq:bch1}
\end{align}
where $\calJ^{-1}_{\boxi}$ is the 9$\times$9 inverse left-Jacobian of $SE_2(3)$.  If both $\boxi$ and $\boeta$ are small quantities, we recover the first-order approximation
\begin{align}
    \log\big(\exp\left(\boxi\right)\exp\left(\boeta\right)\big)    &= \boxi + \boeta + O(\|\boxi\|^2,\|\boeta\|^2).\label{eq:bch2}
\end{align}

Closed-form expressions of $\exp(\cdot)$, $\log(\cdot)$ and Jacobians are given in Appendix.
\begin{rmk}[Overloading Operators]
    We overload the $\exp(\cdot)$, $\log(\cdot)$, and Jacobian in the sense that they can be applied to both 3$\times$1 and 9$\times$1 vectors for respectively $SO(3)$ and $SE_2(3)$ exponential, logarithm and Jacobian.
\end{rmk}

\subsection{Uncertainty \& Random Variables on $SE_2(3)$}\label{sec:uncertainty}
Let $\hbfT \in SE_2(3)$ represent a noise-free value, and $\boxi \in \bbR^9$ a small perturbation. The approach based on additive uncertainty ``$\hbfT + \boxi$'' is not valid as these quantities are not vector elements. In contrast, we define a random variable on $SE_2(3)$ 
\begin{align}
    \bfT := \hbfT \exp(\boxi), \label{eq:bfT}
\end{align}
where $\hbfT$ is a noise-free “mean” of the distribution and $\boxi \sim \calN(\bfzero_9, \boSigma)$ is a zero-mean multivariate Gaussian in $\bbR^9$. We thus describe statistical dispersion of extended poses with the exponential map. This approach is referred to as ``concentrated Gaussian" \cite{bourmaud_continuous_2015,wang_error_2006}, and has been largely advocated for Lie groups, see \cite{barrau_invariant_2018,barrau_invariant_2017,brossard_unscented_2017,barfoot_associating_2014}. 
\begin{rmk}[Left and Right Perturbations]
    We apply perturbations in \eqref{eq:bfT} on the right rather than the left as in \cite{barfoot_associating_2014}. Both are valid, right perturbations are more convenient to propagate extended pose uncertainties in the context of this paper, and one can move from right perturbations to left perturbations and vice-versa through the relation \eqref{eq:adjoint2}.
\end{rmk}
\section{IMU Equations on Flat Earth Revisited}\label{sec:IMU}
This section presents the IMU kinematic model on flat Earth (e.g. with an  inertial  frame  of  reference) and transposes it into the form  \eqref{eq:com_ext_pos}. This was introduced in the bias- and noise-free case in \cite{barrau_linear_2019}, see proofs therein. Noises, biases and rotating Earth are progressively addressed in the following sections.

\subsection{IMU Kinematic Equations on Flat Earth}
Let $\bfR$ denote the rotation matrix encoding the orientation of the IMU, i.e. the rotation from the global frame to the local inertial frame, and let $\bfv$ and $\bfp$ denote the velocity and the position of the IMU expressed in the global frame. An IMU collects angular velocity $\boomega^m$ and proper acceleration $\bfa^m$ measurements which relate to the corresponding true values $\boomega$ and $\bfa$ as
\begin{align}
    \boomega^m &= \boomega + \bfb^{\boomega} + \boeta^{\boomega}, \label{eq:gyro} \\
    \bfa^m &= \bfa + \bfb^{\bfa} + \boeta^{\bfa}.\label{eq:acc}
\end{align}    
The measurements are corrupted both by the time-varying biases $\bfb^{\boomega}$ and $\bfb^{\bfa}$, and zero-mean white Gaussian noises $\boeta^{\boomega}$ and $\boeta^{\bfa}$.  The motion equations of the IMU sensor on flat Earth write
\begin{align}
    \dbfR &= \bfR \left(\boomega^m - \bfb^{\boomega} - \boeta^{\boomega}\right)_\times, \label{eq:dbfR} \\
    \dbfv &= \bfR \left(\bfa^m-\bfb^{\bfa}-\boeta^{\bfa}\right) + \bfg,\label{eq:dbfv} \\
    \dbfp &= \bfv, \label{eq:dbfp}
\end{align}
\enbleu{where $\bfg=[0~ 0~ -9.81]^T \in \bbR^3$ is the global gravity vector.}

\subsection{Revisiting IMU Equations for Extended Poses}

Following \cite{barrau_linear_2019}, we may rewrite the model \eqref{eq:dbfR}-\eqref{eq:dbfp} in the  form   \eqref{eq:com_ext_pos} as follows. First, we associate a matrix $\bfT_{t} \in SE_2(3)$ to the extended pose $\{\bfR, \bfv, \bfp\}$ at time $t$. Then, we write $\bfT_0$ the solution initialized at $t_0=0$, such that
\begin{align}
    \bfT_{t} = \boGamma_{t} \Phi_{t}(\bfT_0) \boUpsilon_{t}, \label{eq:modelt}
\end{align}
where $\Phi_{t}(\cdot)$ and $\boGamma_t$ only depend  on $t$ as
\begin{align}
    \Phi_{t}(\bfT) &:= \begin{bmatrix}\begin{array}{c|cc}
        \bfR & \multicolumn{1}{c|}{\bfv} & \bfp + t \bfv \\ \midrule
        \bfzero_{2\times3} & \multicolumn{2}{c}{\bfI_2}
    \end{array}\end{bmatrix},\label{eq:defPhi} \\
    \boGamma_{t} &:= \begin{bmatrix}\begin{array}{c|cc}
        \bfI_3 & \multicolumn{1}{c|}{t\bfg} & \bfg t^2/2 \\ \midrule
        \bfzero_{2\times3} & \multicolumn{2}{c}{\bfI_2}
    \end{array}\end{bmatrix}, \label{eq:defGamma}
\end{align}
and where $\boUpsilon_{t}$ is solution to differential equations leading to
\begin{align}
    \boUpsilon_{t} := \begin{bmatrix}\begin{array}{c|cc}
        \Delta \bfR_{t} & \Delta \bfv_{t} & \Delta \bfp_{t}\\ \midrule
        \bfzero_{2\times3} & \multicolumn{2}{c}{\bfI_2}
    \end{array}\end{bmatrix},\label{eq:Upsilon}
\end{align}
where
\begin{align}
\Delta \bfR_{t} &=\exp\bigg( \int_{0}^t \boomega dt\bigg)= \exp\bigg(\int_{0}^t \left(\boomega^m - \bfb^{\boomega} - \boeta^{\boomega}\right) dt\bigg),  \label{eq:DeltaR}\\
\Delta \bfv_{t} &= \int_{0}^t \Delta \bfR_{t} \bfa dt = \int_{0}^t \Delta \bfR_{t} \left(\bfa^m - \bfb^{\bfa} - \boeta^{\bfa}\right) dt,\\
\Delta \bfp_{t} &= \int_{0}^t \Delta \bfv_{t}  dt.\label{eq:Deltap}
\end{align}
\enbleu{The reader can check differentiating the product \eqref{eq:modelt} and using \eqref{eq:defPhi}-\eqref{eq:Upsilon} that we recover \eqref{eq:dbfR}-\eqref{eq:dbfp}. } 
$\Delta\bfR_{t}$, $\Delta\bfv_{t}$ and $\Delta\bfp_{t}$, referred to as the  preintegrated measurements in \cite{forster_-manifold_2017}, are quantities based solely on the inertial measurements and do not depend on the initial state $\bfT_0$. This allows to define a unique factor between extended poses at arbitrary temporally distant keyframes based on a unique integration of IMU outputs. 

We now consider discrete time steps with time interval $\Delta t$. Denoting $\boGamma_{i} := \boGamma_{\Delta t}$, $\Phi := \Phi_{\Delta t}$, and $\boUpsilon_{i} := \boUpsilon_{\Delta t}$ we get
\begin{align}
    \bfT_{i+1}=\boGamma_i\Phi(\bfT_{i})\boUpsilon_i, \label{eq:com_ext_pos2}
\end{align}
where $\boGamma_i$, $\bfT_i$, and $\boUpsilon_i$ all live in $SE_2(3)$. The reader may readily check that $\Phi(\boGamma\bfT) = \Phi(\boGamma)\Phi(\bfT)$ and $\Phi\left(\exp\left(\boxi\right)\right) = \exp(\bfF\boxi)$, that we combine as
\begin{align}
    \Phi\big(\bfT\exp(\boxi)\big) =  \Phi\left(\bfT\right)\exp\left(\bfF\boxi\right), \label{eq:phiprop}
\end{align}
where
\begin{align}
    \bfF := \bfF_{\Delta t} :=\begin{bmatrix}
        \bfI_3 & \bfzero_{3\times 3} & \bfzero_{3\times 3} \\
        \bfzero_{3\times 3} &\bfI_3 & \bfzero_{3\times 3} \\
        \bfzero_{3\times 3} & \Delta t \bfI_3 & \bfI_3
    \end{bmatrix}.
\end{align}

\begin{rmk}[Exact Discretization]
The formula \eqref{eq:com_ext_pos2} is an exact discretization of \eqref{eq:dbfR}-\eqref{eq:dbfp}. However it involves \eqref{eq:DeltaR}-\eqref{eq:Deltap} that need to be numerically solved at some point. As IMU measurements come in discrete-time at a high rate, we may call $\Delta t$ the discretization step, assume measurements to be constant over time intervals $\Delta t$, and perform Euler, midpoint, or more sophisticated integration schemes \cite{eckenhoff_closed-form_2019,shen_tightly_2015,henawy_accurate_2020,forster_-manifold_2017}.
\end{rmk}

\subsection{Noise Model and Approximation}\label{sec:justification}

An IMU actually measures noisy observations \eqref{eq:gyro}-\eqref{eq:acc} and we consider IMU noise in the form of 
\begin{align}
    \boUpsilon_i:=\hboUpsilon_i\exp(\boeta_i), \quad \boeta_i \sim \calN(\bfzero_9,\bfQ),\label{eq:ups_noise}
\end{align}
where $\boUpsilon_i$ refers to the true quantity and $\hboUpsilon_i=\hboUpsilon_i(\hbfb_i)$ refers to the estimated one that has been computed with estimated biases 
\begin{align}
    \hbfb_i := \begin{bmatrix}
        \hbfb^{\boomega}_i \\  \hbfb^{\bfa}_i
    \end{bmatrix}.
\end{align}
To justify \eqref{eq:ups_noise}, let us integrate IMU kinematics \eqref{eq:dbfR}-\eqref{eq:dbfp} for one time step with constant global acceleration: 
\begin{align}
    \hboUpsilon_i = \begin{bmatrix}\begin{array}{c|cc}
        \exp(\hboomega_i \Delta t) & \hbfa_i \Delta t & \hbfa_i \Delta t^2/2 \\\midrule
        \bfzero_{2\times3} & \multicolumn{2}{c}{\bfI_{2}}
    \end{array}\end{bmatrix}, \label{eq:onestep}
\end{align}
where $\hboomega_i = \boomega^m_i-\hbfb_i^{\boomega}$ and $\hbfa_i = \bfa^m_i-\hbfb_i^{\bfa}$. This leads to the first-order in the noise terms to
\begin{align}
    \boUpsilon_i \equal{eq:bch1} \hboUpsilon_i \exp\Big(\underbrace{\bfG_i \begin{bmatrix}
        \boeta^{\boomega}_i\\\boeta^{\bfa}_i 
    \end{bmatrix}}_{\boeta_i} + O(\|\boeta_i\|^2)\Big),
\end{align}
where we define the 9$\times$6 matrix
\begin{align}
    \bfG_i = -\begin{bmatrix}
        \calJ_{\hboomega_i\Delta t}^{-1}\Delta t & \bfzero_{3\times3} \\
        \bfzero_{3\times3} & \exp(-\hboomega_i \Delta t)\Delta t \\
        \bfzero_{3\times3} & \exp(-\hboomega_i \Delta t)\Delta t^2/2
    \end{bmatrix},\label{eq:G}
\end{align}
such that the zero-mean noise $\boeta_i$ has covariance 
\begin{align}
    \bfQ_i=\bfG_i\cov\left(\begin{bmatrix}
        \boeta^{\boomega}_i\\\boeta^{\bfa}_i
    \end{bmatrix}\right)\bfG_i^T,
\end{align}
\enbleu{where it is not necessary to assume that IMU noises are isotropic.}
\enbleu{We fix $\bfQ_i=\bfQ$ through this paper instead of defining a time varying $\bfQ_i$ for convenience of exposition.}
\begin{rmk}[Bias Error]
    As in \cite{forster_-manifold_2017}, we start assuming in \eqref{eq:ups_noise} that biases are known. We then investigate how an estimated bias update affects $\hboUpsilon_i$ in Section \ref{sec:pre_bias}. 
\end{rmk}

\begin{rmk}[Integration Scheme]
    \enbleu{We assume in \eqref{eq:onestep} piecewise constant global acceleration as in \cite{forster_-manifold_2017} for convenience of exposition. Opting for a different integration scheme, e.g., with constant local acceleration \cite{eckenhoff_closed-form_2019}, would modify the values inside \eqref{eq:onestep} and \eqref{eq:G} and leave the rest of paper unchanged (details are given in the supplementary material). There also exists polynomial structures for higher order uncertainty tracking \cite{farrell_aided_2008}, see and \cite{musoff_fundamentals_2009} for discussions in the Gaussian context.}
    \end{rmk}
\section{Propagating Uncertainty of Extended Poses}\label{sec:propagating}
The goal of the present section is to compute uncertainty of an extended pose given an initial uncertainty and IMU noises.  

\begin{figure}
    \centering
    \includegraphics{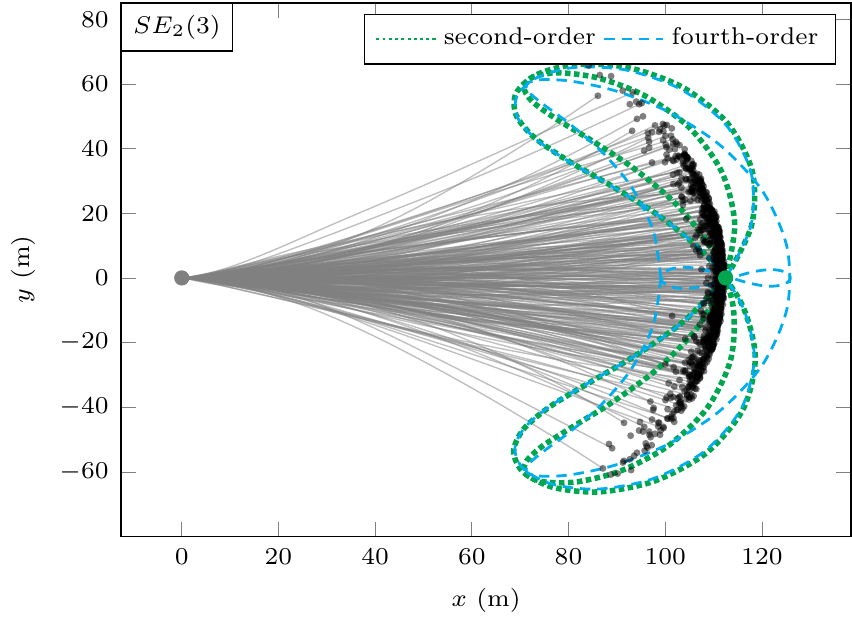}
    \caption{Example of propagation of an extended pose with $SE_2(3)$ uncertainty representation. The green (second-order) and cyan (fourth-order) lines are the principal great circles of the $3\sigma$ covariance ellipsoid, given by $\boSigma_K$, mapped to the $xy$-plane. Looking at the area around the green dot (112.5, 0), corresponding to straight ahead and noise-free model, the fourth-order scheme has some nonzero uncertainty similarly to  the Monte-Carlo samples (black dots), whereas the second-order scheme does not. The small cyan lemniscate around (112.5, 0) corresponds to a great circle of the fourth-order distribution which is absent in the second-order one. \label{fig:uncertainty}}
\end{figure}

\subsection{Propagating Uncertainty through Noise-Free IMU Model}
Let 
\begin{align}
    \bfT_i := \hbfT_i\exp(\boxi_i), \quad \boxi_i \sim \calN(\bfzero_9, \boSigma_i),\label{eq:initnoise}
\end{align}
be our extended pose estimate with error $\boxi_i$. The propagation of $\bfT_i$ through \emph{noise-free} model \eqref{eq:com_ext_pos2} writes
\begin{align}
    \bfT_{i+1} &\equal{eq:com_ext_pos2} \boGamma_i\Phi(\bfT_i)\hboUpsilon_i \equal{eq:initnoise}\boGamma_i\Phi\big(\hbfT_i\exp(\boxi_i)\big)\hboUpsilon_i \nonumber \\
    &\equal{eq:phiprop} \boGamma_i\Phi(\hbfT_i)\Phi\big(\bfF\exp(\boxi_i)\big)\hboUpsilon_i \nonumber\\
     &\equal{eq:adjoint2}\underbrace{\boGamma_i\Phi(\hbfT_i)\hboUpsilon_i}_{\hbfT_{i+1}}\exp(\underbrace{\Ad_{\hboUpsilon_i^{-1}}\bfF \boxi_i}_{\boxi_{i+1}}).\label{eq:mean1}
\end{align}
By defining the new mean $\hbfT_{i+1}$, the covariance of the discrepancy evolves \emph{without} approximation as
\begin{align}
    \boSigma_{i+1}:=\E[\boxi_{i+1}\boxi_{i+1}^T] &\equal{eq:mean1} \underbrace{\Ad_{\hboUpsilon_i^{-1}}\bfF}_{:=\bfA_i} \boSigma_i \left(\Ad_{\hboUpsilon_i^{-1}}\bfF\right)^T.\label{eq:sigmai}
\end{align}
This  is remarkable as it proves that IMU noise-free equations,  albeit nonlinear,  preserve concentrated Gaussians on $SE_2(3)$ and the moments evolve through closed-form formulas. 

\subsection{Propagating Uncertainty through Noisy IMU Model}

We now consider IMU noise in the form of \eqref{eq:ups_noise}. Our Lie group approach to extended poses allows  transposing the developments  of \cite{barfoot_associating_2014} exposed for the evolution model  \eqref{eq:com_pos2}.  The propagation of $\bfT_i$ through \emph{noisy} model writes
\begin{align}
    \bfT_{i+1} = \underbrace{\boGamma_i\Phi(\hbfT_i)\hboUpsilon_i}_{\hbfT_{i+1}}\exp(\underbrace{\Ad_{\hboUpsilon_i^{-1}}\bfF\boxi_i}_{\boxi})\exp\left(\boeta_i\right), \label{eq:mean}
\end{align}
where we recover \eqref{eq:mean1} with noise injected on the right. For our approach to hold, we require that $\E[\boxi_{i+1}] = \bfzero_9$, where
\begin{align}
    \boxi_{i+1} &\equal{eq:initnoise} \log(\hbfT_{i+1}^{-1}\bfT_{i+1})\nonumber\\
    &\equal{eq:mean} \log\big(\exp\left(\boxi\right)\exp\left(\boeta_i\right)\big).
\end{align} 
Applying the BCH formula \eqref{eq:bchapp}, we get
\begin{align}
    \boxi_{i+1}&= \boxi + \boeta_i + \frac{1}{2}\boxi^\curlywedge\boeta_i + \frac{1}{12}\big(
    \boxi^\curlywedge\boxi^\curlywedge\boeta_i +  \boeta_i^\curlywedge\boeta_i^\curlywedge\boxi \big)\nonumber  \\
    &~~ -\frac{1}{24}\boeta_i^\curlywedge\boxi^\curlywedge\boxi^\curlywedge\boeta_i + O(\|\boxi_{i+1}\|^5)
\end{align}
As $\boxi$ and $\boeta_i$ are uncorrelated, we have
\begin{align}
    \E[\boxi_{i+1}] = -\frac{1}{24}\boeta_i^\curlywedge\boxi^\curlywedge\boxi^\curlywedge\boeta_i + O(\|\boxi_{i+1}\|^6) \label{eq:uncornoise}
\end{align}
since everything except the fourth-order term has zero mean. \enbleu{ In the present paper we use ``fourth-order" in the sense of \cite{barfoot_associating_2014} but it is important to note it corresponds to ``second-order" in the original terminology of \cite{yunfeng_wang_nonparametric_2008}. } Thus, to third-order, we can safely assume that $\E[\boxi_{i+1}]=\bfzero_9$, and thus, $\hbfT_{i+1}$ seems to be a reasonable way to compound the mean.  Our goal is now to compute an approximation of $\boSigma_{i+1}$. Multiplying out $\E[\boxi_{i+1}\boxi_{i+1}^T]$ to fourth-order \enbleu{ (second-order in the sense of \cite{yunfeng_wang_nonparametric_2008}), }  the resulting covariance is then
\begin{align}
    \boSigma_{i+1} \simeq \bfA_i \boSigma_{i} \bfA_i^T + \bfQ + \bfS_{\text{4th}}, \label{eq:sigmak1}
\end{align}
where $\bfA_i$ is defined in \eqref{eq:sigmai}, $\bfQ$ is the input noise covariance, and $\bfS_{\text{4th}}$ is the third- and fourth-order corrections to allow \eqref{eq:sigmak1} to be correct to the fourth-order, see Appendix for its expression. This result is essentially the same as \cite{yunfeng_wang_nonparametric_2008,barfoot_associating_2014} but worked out for the time time-varying model \eqref{eq:com_ext_pos2} which is more complex than \eqref{eq:com_pos2}. In summary, to propagate an extended pose, we compute the mean using \eqref{eq:mean} and the covariance using \eqref{eq:sigmak1}.

\subsection{Simple Propagation Example}\label{sec:simple_exm}

This subsection presents a simple example of extended pose propagation. This can be viewed as a discrete-time integration of the IMU kinematic equations \eqref{eq:dbfR}-\eqref{eq:dbfp} with biases absent or perfectly known. To see the qualitative difference between the proposed second- and fourth-order methods, respectively without and with $\bfS_{\text{4th}}$, let us propagate an extended pose many times in a row. We  apply \eqref{eq:com_ext_pos2} for $i=0,\ldots,K-1$ with acceleration about the $x$-axis and rotational noise about the $z$-axis as
\begin{align*}
    \hbfT_0 &= \bfI_5,~~ \boxi_0 \sim \calN(\bfzero_9, \bfzero_{9\times9}), \\
    \cov(\boeta^{\boomega}_i) &= \diag(0,0,\sigma^2),~~~ \cov(\boeta^{\bfa}_i)=\bfzero_{3\times3},\\
    \hboUpsilon_i &= \begin{bmatrix}\begin{array}{c|cc}
        \bfI_3 & \bbfa\Delta t & \bbfa \Delta t^2/2 \\\midrule
        \bfzero_{2\times3}& \multicolumn{2}{c}{\bfI_2}
    \end{array}\end{bmatrix}, 
    \bbfa = \begin{bmatrix}a\\0\\9.81\end{bmatrix}-\bfg.
\end{align*}
This models a robot driving in the plane with a constant translational acceleration and slightly uncertain rotational speed. We are interested in how the covariance matrix fills in over time. According to the second-order scheme, we have
\begin{align*}
    \hbfT_K &= \begin{bmatrix}\begin{array}{c|cc}
        &K a \Delta t& K^2 a^2 \Delta t^2/2\\
        \bfI_3&0&0\\
        &0&0\\\midrule
        \bfzero_3^T&\multicolumn{2}{c}{\bfI_2}
    \end{array}\end{bmatrix}, \\
    \boSigma_K &= \begin{bmatrix}\begin{array}{ccc|ccc|ccc}
        0&0&0&0&0&0&0&0&0\\
        0&0&0&0&0&0&0&0&0\\
        0&0&1&0&\Sigma_{\phi v}&0&0&\Sigma_{\phi p}&0\\\midrule
        0&0&0&0&0&0&0&0&0\\
        0&0&\Sigma_{\phi v}&0&\Sigma_{v v}&0&0&\Sigma_{v p}&0\\
        0&0&0&0&0&0&0&0&0\\\midrule
        0&0&0&0&0&0&\underline{0}&0&0\\
        0&0&\Sigma_{\phi p}&0&\Sigma_{v p}&0&0&\Sigma_{p p}&0\\
        0&0&0&0&0&0&0&0&0
    \end{array}\end{bmatrix} K \sigma^2,
\end{align*}
where
\begin{align*}
    \Sigma_{\phi v} &= -\frac{\left(K-1\right)}{2} a \Delta t,\\
    \Sigma_{\phi p} &= \frac{\left(K-1\right)\left(2K-1\right)}{12} a \Delta t^2,\\
    \Sigma_{v v} &= \frac{\left(K-1\right)\left(2K-1\right)}{6} a^2 \Delta t^2,\\
    \Sigma_{v p} &= \frac{\left(K-1\right)^2 K}{8} a^2 \Delta t^3,\\
    \Sigma_{p p} &= \frac{\left(K-1\right)\left(2K-1\right)\left(3\left(K-1\right)^2+3K-4\right)}{120} a^2 \Delta t^4.
\end{align*}
We see that the entry of $\boSigma_K$ corresponding to uncertainty in the $x$-direction (underlined zero just left above $\Sigma_{p p}$), does not grow. However, in the fourth-order scheme, the fill-in pattern is such that this entry is nonzero. This leaking of uncertainty into an additional degree of freedom cannot be captured by keeping only the second-order terms. Figure \ref{fig:uncertainty} provides a numerical example of this effect, where we set $K=300$, $dt=\SI{0.05}{s}$, $a = \SI{1}{m/s^2}$ and $\sigma = \SI{0.03}{rad/s}$. It shows that:
\begin{enumerate}
    \item both the second- and fourth-order schemes capture the actual ``banana''-shaped distribution over extended poses of the Monte-Carlo samples. This is owed to  the use of the exponential of $SE_2(3)$; 
    \item the fourth-order scheme has some finite uncertainty in the straight-ahead direction (the gray dot to the green dot) similarly to  the sampled trajectories, while the second-order scheme does not; 
    \item the samples never cross a circle of radius \SI{112.5}{m},   as rotation uncertainty tends to reduce the travelled distance, and none of the approximate methods perfectly models this non-Gaussian behavior.
\end{enumerate}

\begin{figure}
    \centering
    \includegraphics{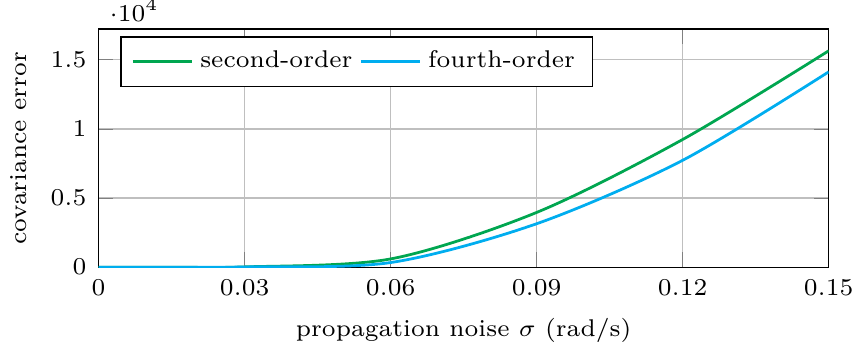}
    \caption{Error in computing covariance using second- and fourth-order methods for propagating an extended pose, as compared to Monte-Carlo. The input noise is gradually scaled up, highlighting the improved performance of the fourth-order method. \label{fig:scale}}
\end{figure}

\begin{figure*}
    \begin{minipage}{.25\textwidth}
    {\centering\underline{$SO(3)\times \bbR^6$} \cite{forster_-manifold_2017}}
    \small
    \begin{align*}
        \bfT &= \hbfT \boxplus \boxi \\
    &= \begin{bmatrix}\begin{array}{c|cc}
        \hbfR\exp(\bophi) & \hbfv +\bonu & \hbfp + \hbfR\borho\\\midrule
        \bfzero_{2\times3} & \multicolumn{2}{c}{\bfI_{2}}
    \end{array}\end{bmatrix} \\
    \boxi &= \bfT \boxminus \hbfT = \begin{bmatrix}
        \log(\hbfR^T\bfR) \\ \bfv- \hbfv \\ \hbfR^T(\bfp-\hbfp)
    \end{bmatrix}
    \end{align*}
    \end{minipage}
    \hfill\vline\hfill
    \begin{minipage}{.25\textwidth}
        {\centering \underline{$SE_2(3)$}}
        \begin{align*}
        \bfT &= \hbfT \exp(\boxi) \\
    &= \begin{bmatrix}\begin{array}{c|c|c}
        \hbfR\exp(\bophi) & \hbfv + \calJ_{\bophi}\bonu & \hbfp + \calJ_{\bophi}\borho\\\midrule
        \bfzero_{2\times3} & \multicolumn{2}{c}{\bfI_{2}}
    \end{array}\end{bmatrix} \\
    \boxi &= \log(\hbfT^{-1}\bfT)
    \end{align*}
    \hfill\vline\hfill
    \end{minipage}
    \hfill\vline\hfill
    \begin{minipage}{.25\textwidth}
        \centering
        \includegraphics{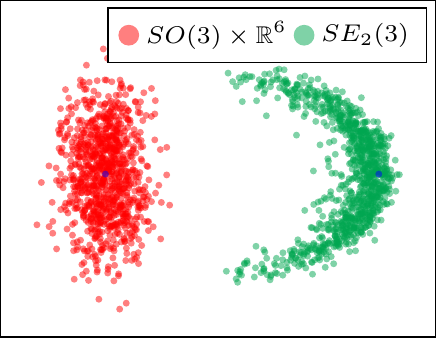}
    \end{minipage}
    \caption{Comparison between the retraction and its inverse used in the preintegration on manifold \cite{forster_-manifold_2017} (left), and the ones based on $SE_2(3)$ exponential advocated in the present paper (center, detailed formulas are given in Appendix). On the right, the $SE_2(3)$ distribution (green dots) is curved whereas the $SO(3)\times\bbR^6$ one (red dots) is limited to  ellipses. The blue points show the mean estimate. \label{fig:retractions}}
    \end{figure*}

\subsubsection{Comparison with $SE_2(3)$ Monte-Carlo Distribution}
assuming the actual distribution to be a concentrated Gaussian on $SE_2(3)$, the Monte-Carlo estimate of its covariance reads
\begin{align*}
    \boSigma_{\mc} &:= \E\left[\log(\hbfT_K^{-1}\bfT_K)\log(\hbfT_K^{-1}\bfT_K)^T\right]\nonumber \\
    &\simeq \frac{1}{N-1} \sum_{n=1}^{N} \log(\hbfT_K^{-1}\bfT_{K,n})\log(\hbfT_K^{-1}\bfT_{K,n})^T.
\end{align*}By letting   $N=\num{e6}$ this provides a benchmark to compare the covariance matrices computed with second-order accuracy and fourth-order accuracy based on the Frobenius norm \cite{barfoot_associating_2014}
\begin{align*}
    \text{cov. err.} := \sqrt{\trace\left(\left(\boSigma_{i\text{-th}}-\boSigma_{\mc}\right)^T\left(\boSigma_{i\text{-th}}-\boSigma_{\mc}\right)\right)}.
\end{align*}
Results are displayed in Figure \ref{fig:scale}. We see the fourth-order is a slightly better. All methods degrade as magnitude of noise increases and fourth-order is the closest to the true covariance.

\begin{figure}
    \centering
    \includegraphics{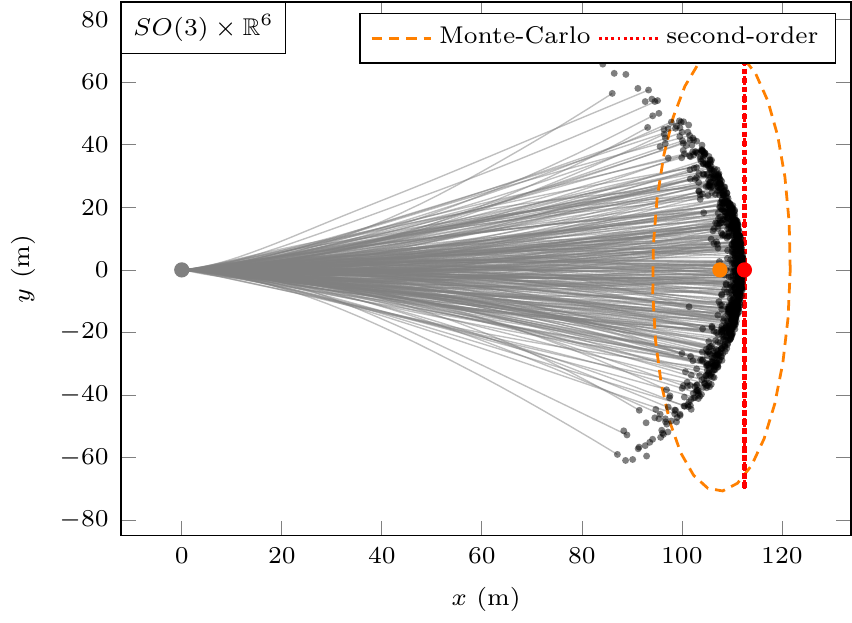}
    \caption{Same example as Figure \ref{fig:uncertainty} adapted to the $SO(3) \times \bbR^6$ uncertainty representation. The mean and the $3\sigma$ covariance ellipse relative to the Monte-Carlo distribution (orange) are fitted to the black samples to show what keeping $xy$-covariance relative to the start looks like. The red line is the $3\sigma$ ellipsoid when incrementally computing the covariance with second-order accuracy, which wrongly reveals null uncertainty in the $x$-direction. The position mean of the samples (orange dot) differs from the noise-free mean (red dot). \label{fig:uncertainty2}}
\end{figure}

\subsubsection{Comparison with $SO(3)\times \bbR^6$ Monte-Carlo Distribution}
we compare the $SE_2(3)$ uncertainty representation to the gold-standard $SO(3)\times \bbR^6$ distribution used in \cite{forster_-manifold_2017} for preintegration, see Section \ref{sec:preintegration}. This distribution defines uncertainty as
\begin{align*}
    \bfT &:= \hbfT \boxplus \boxi, ~~ \boxi \sim \calN(\bfzero_9, \boSigma),
\end{align*}
where the $\boxplus$ retraction substitutes our $SE_2(3)$ exponential map and is defined in Figure \ref{fig:retractions}. This avoids the computation of 3$\times$3 Jacobian but comes at the price of loosing the ability to model the ``banana'' shape of the actual dispersion.

We compute the best mean and covariance related to the $SO(3)\times \bbR^6$ distribution by averaging Monte-Carlo samples 
\begin{align*}
    \hbfT_K &:= \E[\bfT_K] := \begin{bmatrix}\begin{array}{c|cc}
        \E[\bfR_K] & \E[\bfv_K] & \E[\bfp_K]\\\midrule
        \bfzero_{2\times3}&\multicolumn{2}{c}{\bfI_2}
    \end{array}\end{bmatrix}, \\
    \boSigma &:= \E\left[\big(\bfT_K\boxminus \hbfT_K\big)\big(\bfT_K\boxminus \hbfT_K\big)^T\right] \\
    &\simeq \frac{1}{N-1} \sum_{n=1}^{N} \big(\bfT_{K,n}\boxminus \hbfT_K\big)\big(\bfT_{K,n}\boxminus \hbfT_K\big)^T, 
\end{align*}
where $\E[\bfR_K] \simeq 1/N \exp(\sum_{n=1}^N \log(\hbfR^n_K))$, $\E[\bfv_K]\simeq 1/N \sum_{n=1}^N \hbfv^n_K$ and $\E[\bfp_K] \simeq 1/N \sum_{n=1}^N \hbfp^n_K$. Both approaches are compared in Figure \ref{fig:uncertainty2}. We see that 
\begin{enumerate}
    \item the covariance ellipse (orange line) is not a good fit for samples with large deviation as compared to the $SE_2(3)$ distribution for the same experiment;
    \item the mean of the samples (orange dot) is distant from  \SI{5}{m} of the noise-free estimate (red dot).
\end{enumerate}
Let us explain why our $SE_2(3)$-based method improves on point  2).  Consider the position extracted from the extended pose
\begin{align}
    \bfp_K &= \bfT_K \begin{bmatrix}\bfzero_4 \\ 1\end{bmatrix}\equal{eq:initnoise} \hbfT_K \exp(\boxi_K) \begin{bmatrix}\bfzero_4 \\ 1\end{bmatrix} \nonumber\\
    &\equal{eq:bchapp} \hbfp_K + \borho + \frac{1}{2} \bophi_\times \borho + O(\|\boxi_K\|^3). \label{eq:pK}
\end{align}
where $\hbfp_K= [\SI{112.5}{m}, \SI{0}{m}, \SI{0}{m}]$ up to third-order, see \eqref{eq:uncornoise}. Let us  compute the expectation of the position as 
\begin{align}
    \E[\bfp_K] \equal{eq:pK} \hbfp_K + \underbrace{\E[\borho]}_{=\bfzero_3} + \frac{1}{2} \underbrace{\E[\bophi_\times \borho]}_{\neq \bfzero_3} + O(\|\boxi_K\|^3) \neq \hbfp_K, \label{eq:meanpK}
\end{align}
where we see a shift appear due to the correlation between orientation and position. Using the values of the covariance $\boSigma_K$ of the $SE_2(3)$ second-order distribution, we find $\E[\bfp_K]= [\SI{107.5}{m}, \SI{0}{m}, \SI{0}{m}]$ which matches with the Monte-Carlo mean value (orange dot) of Figure \ref{fig:uncertainty2}.

\subsubsection{Comparison with $SO(3)\times \bbR^6$ $2^\text{nd}$-Order Distribution}
in practice, e.g., in an extended Kalman filter, the covariance $\boSigma$ is recursively computed  with second-order accuracy, i.e. similarly as \eqref{eq:sigmak1} with $\bfS_{\mathrm{4th}}=\bfzero_{9\times9}$. We compute the covariance $\boSigma$ with the second-order accuracy adapted to the $SO(3)\times \bbR^6$ uncertainty representation. The covariance reflects \emph{null} uncertainty along the $x$-axis due to linearization error, see the red  ``ellipse" in Figure \ref{fig:uncertainty2}, and see also Figure \ref{fig:intro} where accelerometer noises are considered. We provide quantitative comparisons between $SO(3)\times \bbR^6$ and $SE_2(3)$ distributions in the context of preintegration in Section \ref{sec:num_ex}.
%

\subsubsection{Summary} given its apparent simplicity, the latter instructive simulation evidences the $SE_2(3)$ distribution is especially suited to extended pose uncertainty in the two following situations:
\begin{enumerate}
    \item when initial uncertainty predominates input noise uncertainty, as the method is then exact, see formula  \eqref{eq:sigmai};
    \item when orientation uncertainty dominates translational uncertainty, leading to a ``banana''-shaped dispersion.
\end{enumerate}
An interesting question is whether to choose   the noise-free mean $\hbfp_K$ or the ``stochastic'' one \eqref{eq:meanpK} for use in sensor-fusion or  outlier detection algorithms based on the $SO(3)\times \bbR^6$ distribution, e.g. \cite{Chng_outlier-robust_2019}. Yet, it is left for future work.

\subsection{Batch \& Incremental Extended Pose Propagation}
\label{sec:batch}
The previous subsections provide incremental expressions while we infer here batch expressions to compute the mean and the uncertainty of the extended pose 
\begin{align}
    \bfT_j = \boGamma_{ij} \Phi_{ij}(\bfT_i)\boUpsilon_{ij}, \label{eq:extentedj}
\end{align}
 after the integration between two arbitrary instants $t_i$ and $t_j= t_i+(j-i)\Delta t$, which is helpful for preintegration theory, see Section \ref{sec:preintegration}. Propagating an extended pose through model \eqref{eq:com_ext_pos2} between two consecutive time steps, we obtain
\begin{align}
\bfT_{i+2} &\equal{eq:com_ext_pos2} \boGamma_{i+1}\Phi\left(\bfT_{i+1}\right)\boUpsilon_{i+1}  \nonumber\\
&\equal{eq:com_ext_pos2} \boGamma_{i+1}\Phi\big(\boGamma_i\Phi\left(\bfT_{i}\right)\boUpsilon_i\big)\boUpsilon_{i+1}  \nonumber\\
 &\equal{eq:phiprop} \underbrace{\boGamma_{i+1}\Phi\left(\boGamma_{i}\right)}_{\boGamma_{i(i+2)}}\underbrace{\Phi_{2\Delta t}\left(\bfT_i\right)}_{\Phi_{i(i+2)}\left(\bfT_i\right)}\underbrace{\Phi\left(\boUpsilon_i\right)\boUpsilon_{i+1}}_{\boUpsilon_{i(i+2)}}\label{eq:2i}.
\end{align}
Based on \eqref{eq:mean} and \eqref{eq:2i}, a recursion provides the following batch and incremental formulas
\begin{align}
    \hboUpsilon_{ij} &=  \prod_{k=i}^{j-1} \Phi_{k(j-1)}\big(\hboUpsilon_{k}\big) = \Phi\big(\hboUpsilon_{i(j-1)}\big)\hboUpsilon_{j}, \label{eq:ups_batch}\\
    \boGamma_{ij} & =\prod_{k=1}^{j-i} \Phi_{(j-k)(j-1)}\left(\boGamma_{j-k}\right)=  \boGamma_{i(j-1)}\Phi\left(\boGamma_{j}\right),\\
    \boxi_{i+1} &= \log\big(\exp(\bfA_{i}\boxi_{i})\exp(\boeta_i)\big) \nonumber \\
    &= \log\bigg(\exp\left(\bfA_0^{i}\boxi_0\right)\prod_{k=0}^{i}\exp\Big( \underbrace{\prod_{l=k+1}^{i} \bfA_l}_{:=\bfA_{k+1}^{i}}\boeta_k\Big)\bigg)\label{eq:dis}\\
    &\simeqbis{\ref{eq:bch2}} \bfA_{i}\boxi_{i} + \boeta_i = \bfA_0^{i}\boxi_0 + \sum_{k=0}^{i} \bfA_{k+1}^{i}\boeta_k, \\
    \boSigma_{i-1} &\simeq \bfA_0^{i} \boSigma_{0}  \bfA_0^{iT} + \sum_{k=0}^{i} \bfA_{k+1}^{i} \bfQ \bfA_{k+1}^{iT},
\end{align}
which coincide with the continuous model \eqref{eq:dbfR}-\eqref{eq:dbfp} \emph{without} approximation, e.g. we recover 
\begin{align*}
    \boPhi_{ij}(\cdot) \equal{eq:defPhi} \boPhi_{(j-i)\Delta t}(\cdot), ~~  \boGamma_{ij}\equal{eq:defGamma} \boGamma_{(j-i)\Delta t}.
\end{align*}
Remarkably, we have an exact formula  for the evolution of the discrepancy \eqref{eq:dis}, which is rare in nonlinear estimation.

\section{On Lie Group IMU Preintegration}\label{sec:preintegration}

This section first evidences the relations between the IMU preintegration on manifold \cite{forster_-manifold_2017} and our approach on the Lie group $SE_2(3)$. Then we describe how to compute the preintegrated factor along with its noise covariance matrix on $SE_2(3)$. Finally we address incorporation of bias updates in Lie exponential coordinates.

\subsection{Links with the IMU Preintegration on Manifold \cite{forster_-manifold_2017}}

While \eqref{eq:mean} could be readily seen as a probabilistic constraint in a factor-graph, it would require to include states in a factor-graph at the high IMU rate. It indeed relates states at time $t_i$ and $t_{i+1} = t_i + \Delta t$, where $\Delta t$ is the sampling period of the IMU. Lupton and Sukkarieh \cite{lupton_visual-inertial-aided_2012} and then Forster et. al. \cite{forster_-manifold_2017} show that all measurements between two chosen instants $t_i$ and $t_j$ can be summarized in a single factor, which constrains the motion between times $t_i$ and $t_j$.

Our approach differs from \cite{forster_-manifold_2017} in three ways: $i$) the situation where the mathematical framework applies; $ii$) the retraction used to compute Gaussian uncertainty; and $iii$) how bias updates modify the preintegrated factor. Regarding $i$), \cite{forster_-manifold_2017} addresses preintegration by computing factors between two arbitrary timestamps $t_i$ and $t_j$ as
\begin{align}
    \Delta \bfR_{ij} &= \bfR_i^T \bfR_j, \label{eq:drij}\\
    \Delta \bfv_{ij} &= \bfR_i^T\left(\bfv_j-\bfv_i - \bfg \Delta t_{ij}\right),\\
    \Delta \bfp_{ij} &= \bfR_i^T\big(\bfp_j-\bfp_i - \bfv_i \Delta t_{ij} - \frac{1}{2} \bfg \Delta t_{ij}^2\big),\label{eq:prij}
\end{align}
where $\Delta t_{ij}:= t_j-t_i=(j-i)\Delta t$, which is specific to the model \eqref{eq:dbfR}-\eqref{eq:dbfp}. Our preintegration factor between two extended poses is given as
\begin{align}
    \boUpsilon_{ij} &\equal{eq:extentedj}  \Big(\boGamma_{ij} \Phi_{ij}\left(\bfT_i\right)\Big)^{-1}\bfT_{j}, \label{eq:preint_mes} 
\end{align}
and is derived for flat Earth model in this section, but may be adapted to rotating Earth model, see  Section \ref{sec:coriolis}.

The difference $ii$) is the retraction function used to represent uncertainty. While \cite{forster_-manifold_2017} opts for the $SO(3)\times \bbR^6$ uncertainty 
\begin{align}
    \bfT &:= \hbfT \boxplus \boxi, ~~ \boxi \sim \calN(\bfzero_9, \boSigma),\label{eq:stdunc}
\end{align}
where the $\boxplus$ operator is defined in Figure \ref{fig:retractions}, we advocate  the uncertainty based on the $SE_2(3)$ exponential map, see Section \ref{sec:uncertainty}, which proves better suited to propagate extended pose uncertainty (see Section \ref{sec:simple_exm}).
\begin{rmk}
In the implementation of \cite{forster_-manifold_2017} on GTSAM \cite{dellaert_factor_2012}, the retraction updates the velocity in the navigation frame. Since the only difference is the use of an invertible matrix, this is in fact equivalent. We thus do not distinguish both retractions.
\end{rmk}

Finally, given a bias update $\hbfb^+_i \leftarrow \hbfb_i + \delta \bfb$, \cite{forster_-manifold_2017} addresses $iii$) by updating the preintegrated measurements using a first-order Taylor expansion as
\begin{align*}
    \Delta \hbfR_{ij}(\hbfb_i^+) &= \Delta \hbfR_{ij}(\hbfb_i)\exp\Big(\frac{\partial \Delta \hbfR_{ij}}{\partial \bfb}|_{\hbfb_i} \delta \bfb + O(\|\delta\bfb\|^2)\Big), \\
    \Delta \hbfv_{ij}(\hbfb_i^+) &= \Delta \hbfv_{ij}(\hbfb_i) + \frac{\partial \Delta \hbfv_{ij}}{\partial \bfb}|_{\hbfb_i} \delta\bfb+ O(\|\delta\bfb\|^2),\\
    \Delta \hbfp_{ij}(\hbfb_i^+) &= \Delta \hbfp_{ij}(\hbfb_i) + \frac{\partial \Delta \hbfp_{ij}}{\partial \bfb}|_{\hbfb_i} \delta\bfb + O(\|\delta\bfb\|^2).
\end{align*}
To be consistent with our framework, we address the bias update as a first-order Taylor expansion in the Lie exponential coordinates as
\begin{align*}
    \hboUpsilon_{ij}(\hbfb_i^+) = \hboUpsilon_{ij}(\hbfb_i)\exp\Big(\frac{\partial \hboUpsilon_{ij}}{\partial \bfb}|_{\hbfb_i}\delta\bfb+ O(\|\delta\bfb\|^2)\Big).
\end{align*}

\begin{rmk}
    The implementation of \cite{forster_-manifold_2017} in GTSAM \cite{dellaert_factor_2012} takes into account the uncertainty in the bias  estimates for preintegration, such that the preintegration covariance matrix preserves the correlation between the bias uncertainty and the preintegrated measurements uncertainty. \enbleu{We follow \cite{forster_-manifold_2017} for clarity of exposition. However our approach is fully compatible with accounting for the correlation between IMU navigation state and biases states, as is detailed in the supplementary material.}
\end{rmk}

\subsection{Proposed IMU Preintegration}
As in \cite{forster_-manifold_2017}, we first assume the bias estimates are exact and fixed between $t_i$ and $t_j$ as
\begin{align}
    \hbfb_i = \hbfb_{i+1} = \cdots =\hbfb_{j-1}.
\end{align}
\subsubsection{Preintegrating IMU Measurements}
we relate the states at times $t_i$ and $t_j$ to the IMU measurement through 
\begin{align}
    \boUpsilon_{ij} = \hboUpsilon_{ij} \exp(\boeta_{ij}) \equal{eq:preint_mes}  \Big(\boGamma_{ij} \Phi_{ij}\left(\bfT_i\right)\Big)^{-1}\bfT_{j},\label{eq:ups_pre}
\end{align}
which provides a measurement model where the noise terms of the individual inertial measurements is isolated in $\boeta_{ij}$ and where $\hboUpsilon_{ij}$ is integrated with inertial measurements through \eqref{eq:ups_batch}, which can be substituted by the integration \eqref{eq:DeltaR}-\eqref{eq:Deltap}. Indeed, our approach has the same preintegration \emph{measurement} as \cite{forster_-manifold_2017} but the \emph{uncertainty} is encoded in Lie exponential coordinates. 

\subsubsection{Noise Propagation}
we derive the statistics of the noise vector $\boeta_{ij}\sim \calN(\bfzero_9,\boSigma_{ij})$. Following Section \ref{sec:propagating}, $\boeta_{ij}$ is zero-mean up to the third-order. To model the noise covariance, we develop the batch expression as follows
\begin{align}
    \boUpsilon_{ij} &\equal{eq:ups_batch} \prod_{k=i}^{j-1} \Phi_{k(j-1)}(\boUpsilon_{k}) \equal{eq:ups_noise} \prod_{k=i}^{j-1}  \Phi_{k(j-1)}\left(\hboUpsilon_{k} \exp(\boeta_k)\right) \nonumber\\
    &\equal{eq:adjoint2} \underbrace{\prod_{k=i}^{j-1} \Phi_{k(j-1)}\big(\hboUpsilon_{k}\big)}_{\equal{eq:ups_batch}   \hboUpsilon_{ij}}  \prod_{k=i}^{j-1} \exp\Big(\underbrace{\prod_{l=k+1}^{j-1} \Ad_{\hboUpsilon_{l}^{-1}}\bfF}_{\equal{eq:dis}\bfA_{k+1}^{j-1}}\boeta_k\Big) \nonumber \\
    &\equal{eq:bch2} \hboUpsilon_{ij} \exp\bigg(\sum_{k=i}^{j-1}\bfA_{k+1}^{j-1}\boeta_k + O(\|\boeta_k\|^2)\bigg),
\end{align}
that corresponds to integrating the uncertainty of an extended pose without initial uncertainty. We resort to  \eqref{eq:sigmak1} to compute the covariance with second-order accuracy as
\begin{align}
    \boSigma_{i(j+1)} = \Ad_{\hboUpsilon_{j}^{-1}}\bfF \boSigma_{ij} \left(\Ad_{\hboUpsilon_{j}^{-1}}\bfF\right)^T + \bfQ,\label{eq:Sigma}
\end{align}
and starting from initial condition $\boSigma_{ii} = \bfzero_{9\times9}$.

\begin{figure}
    \centering
    \includegraphics{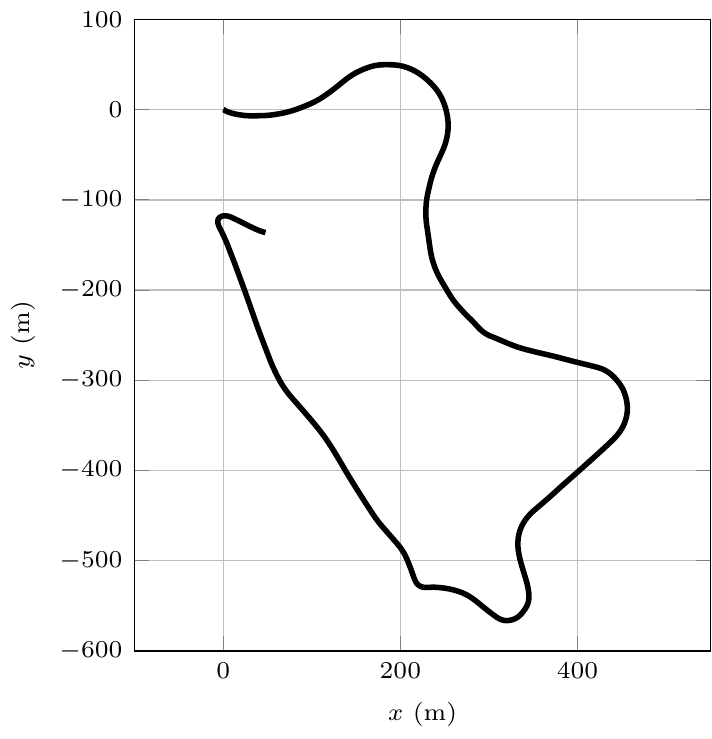}
    \caption{Reference trajectory used during the simulation experiments. It corresponds to a \SI{1.7}{km} car trajectory of the KITTI dataset \cite{geiger_vision_2013} whose duration is \SI{166}{s}.\label{fig:trajsimu}}
\end{figure}
\subsection{Numerical Example without Biases}\label{sec:num_ex}

\enbleu{To quantitatively assess our preintegration technique against state-of-the-art based on closed-form integration scheme \cite{eckenhoff_closed-form_2019} and uncertainty representation \cite{forster_-manifold_2017}, we run a numerical experiment where the IMU follows the  car trajectory of Figure \ref{fig:trajsimu}.} We gather the path of the sequence 9 of the KITTI dataset \cite{geiger_vision_2013}, from which we differentiate ground-truth and infer true inertial measurements at $\Delta t = \SI{0.1}{s}$. The ground-truth is then recomputed with model \eqref{eq:dbfR}-\eqref{eq:dbfp} to remove integration errors that occur when assuming constant measurements in \eqref{eq:DeltaR}-\eqref{eq:Deltap}. We then add Gaussian noise whose covariance is defined as
\begin{align*}
    \cov(\boeta^{\boomega}_i) &= \alpha\sigma_{\boomega}^2 \bfI_3,~~~ \cov(\boeta^{\bfa}_i) = \alpha\sigma_{\bfa}^2 \bfI_3,
\end{align*}
where $\sigma_{\boomega}$ and $\sigma_{\bfa}$ are the same as in simulations of \cite{forster_-manifold_2017}, i.e. \SI{7e-4}{rad/(s$\sqrt{\text{Hz}}$)} and \SI{1.9e-2}{m/(s^2$\sqrt{\text{Hz}}$)},  and with $\alpha$ a scaling parameter for testing purposes.

The factor covariance is computed using four methods:
\begin{enumerate}
    \item $SE_2(3)$, which computes the covariance based on \eqref{eq:Sigma} with accuracy up to second-order;
    \item \enbleu{$SO(3)\times \bbR^6$ \cite{forster_-manifold_2017, eckenhoff_closed-form_2019}, same as the method 1) but adapted to uncertainty on $SO(3)\times \bbR^6$, i.e. with uncertainty representation of \cite{forster_-manifold_2017} and integration scheme of \cite{eckenhoff_closed-form_2019};}
    \item \enbleu{$SE(3)\times \bbR^3$, same as the method 1) but adapted to   uncertainty on $SE(3)\times \bbR^3$ (this method has not appeared before to our knowledge);}
    \item \textbf{$SE_2(3)$ (Monte-Carlo)}, which is a slow yet accurate approach. At each time step, we draw a large number $M=\num{e6}$ of random samples $\boxi^m_{ij}$ and $\boeta_{j}^m$, propagate the resulting states, and compute the covariance as $\boSigma_{i(j+1)} = \frac{1}{M-1}\sum_{m=1}^M \boxi^m\boxi^{mT}$ with 
    \begin{align*}
        \bfT^{m} = \boGamma_j \Phi\left(\hbfT_{ij} \exp(\boxi^m_{ij})\right) \hboUpsilon_{j} \exp(\boeta^m_{j}),
    \end{align*}
    and $\boxi^m = \log(\hbfT_{i(j+1)}^{-1} \bfT^{m})$. It requires, e.g., $50M=\num{5e7}$ samples for one preintegration factor of length $\Delta t_{ij} = \SI{5}{s}$. It indicates  the best recursive distribution based on $SE_2(3)$ one may obtain. 
\end{enumerate}

\begin{figure}
    \centering
    \includegraphics{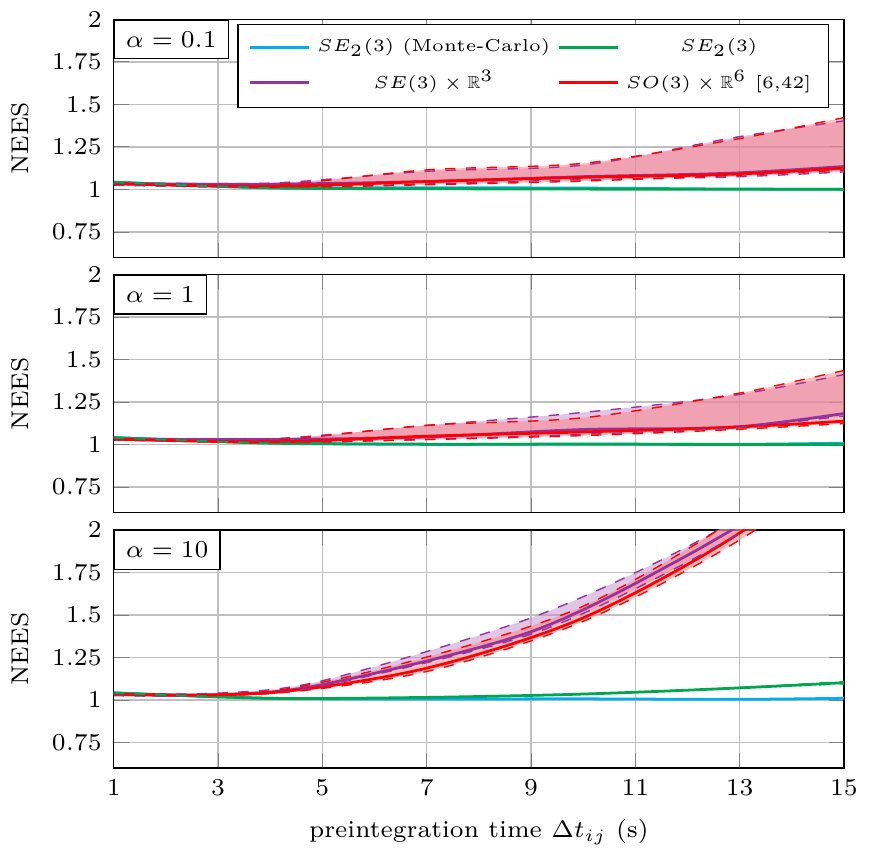}
    \caption{IMU preintegration experiment. NEES criterion based on Monte-Carlo simulations for low noise (above), medium noise (middle), and high noise (bottom). Target value is one, solid curves are the median value, and dashed curves are the 33\% and 67\% percentiles of the errors. Solid and dashed curves coincide for the $SE_2(3)$ methods. The improved performances of the $SE_2(3)$ methods are highlighted as the preintegration time gradually scales up. \label{fig:kl}}
\end{figure}

To evidence the fact that our approach is more consistent, we compute the average Normalized Estimation Error Squared (NEES) as
\begin{align*}
    \text{NEES} = \frac{1}{9N} \sum_{n=1}^N \bfe_n^T \boSigma^{-1} \bfe_n,
\end{align*}
where the error is defined as
\begin{align*}
    \bfe_n^{SE_2(3)} =\log\big(\hbfT_{ij}^{-1} \hbfT_{ij}^{n}\big), \text{~~or~~}
    \bfe_n^{SO(3)\times \bbR^6} = \bfT_{ij}^{n} \boxminus \hbfT_{ij}.
\end{align*}
We sample $N=\num{e6}$ Monte-Carlo noisy preintegrations $\bfT^n_{ij}$ for each factor $\bfT_{ij}$, where for each Monte-Carlo realization $n$, we compute one realization of each noise $\boeta_{i}$, \ldots, $\boeta_{j-1}$. If the NEES is higher than 1, the approach is overconfident, if it stays below 1, it is conservative. One usually wants to avoid overconfident estimates.

Results are displayed in Figure \ref{fig:kl} for three noise scale parameters and increasing values of the preintegration time. We observe that:
\begin{enumerate}
    \item $SE_2(3)$ distribution is the most consistent approach. This becomes   visible as the preintegration time increases for each level of IMU noises;
    \item \enbleu{$SO(3) \times \bbR^6$ distribution poorly approximates long preintegration uncertainty, where the median value is close to the 33\% percentile while the 67\% percentile is much larger. It indicates that there exists some types of trajectory (that correspond to high rotation motions) where the performances of the $SO(3)\times \bbR^6$ on manifold method might degrade. This confirms the results of Section \ref{sec:simple_exm} where we see the average error  should be not zero and that linearization errors do occur;}
    \item \enbleu{$SE(3) \times \bbR^3$ obtains similar results as $SO(3) \times \bbR^6$, and we see that despite the use of $SE(3)$ which is known to describe well uncertainty associated with poses, the full $SE_2(3)$ structure is desirable in the context of IMUs;}
    \item all methods degrade as  noise and preintegration time increase because the preintegrated measurements become non-Gaussian. Interestingly, the linearization errors of our  $SE_2(3)$-based scheme (as opposed to Monte-Carlo estimate) are only visible for high noise ($\alpha=10$) and sufficiently high preintegration time.  
\end{enumerate}  

\subsection{Bias Correction via Lie Exponential First-Order Updates}\label{sec:pre_bias}

This subsection shows the impact of computing first-order bias corrections using the representation of errors based on exponential coordinates of $SE_2(3)$. In the context of preintegration, given a bias update $\hbfb^+_i \leftarrow \hbfb_i + \delta\bfb$, one needs to compute how the preintegrated quantities $\hboUpsilon_{ij}$ change. Assume we have computed $\hboUpsilon_{ij}(\hbfb_i)$  corresponding to bias $\hbfb_i$ and let $\hboUpsilon_{ij}(\hbfb_i^+)$ denote the measurement associated to new bias estimation. We define the first-order update as
\begin{align}
    \hboUpsilon_{ij}(\hbfb_i^+) = \hboUpsilon_{ij}(\hbfb_i)\exp\Big(\frac{\partial \hboUpsilon_{ij}}{\partial \bfb}|_{\hbfb_i}\delta\bfb + O(\|\delta\bfb\|^2)\Big).\label{eq:biasupdate}
\end{align}

\begin{figure}
    \centering
    \includegraphics{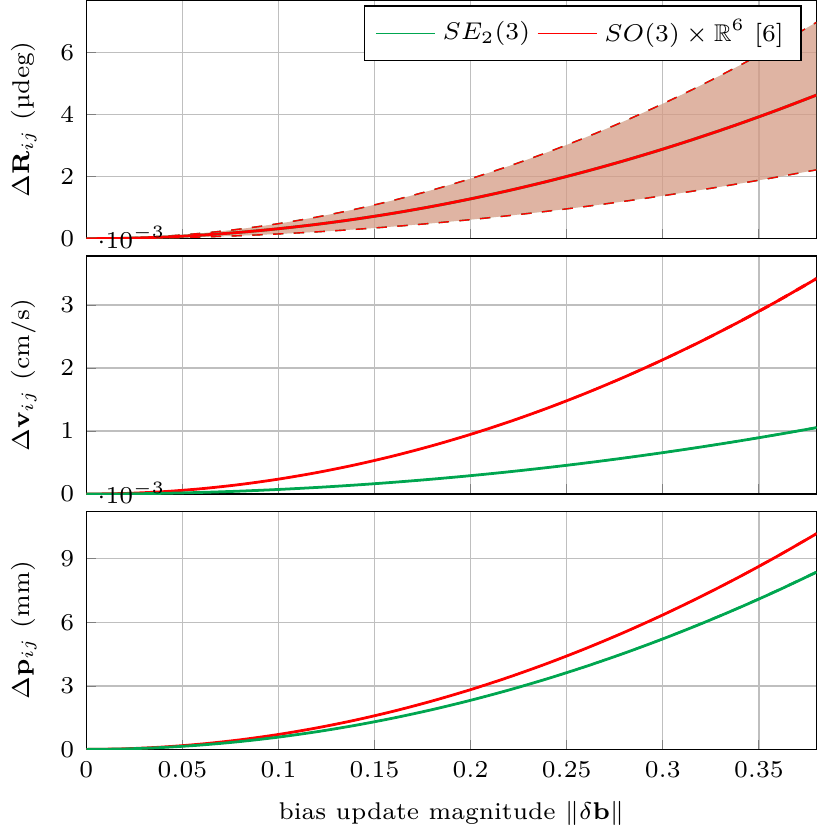}
    \caption{Error committed when using the first-order approximation
    instead of repeating the integration, for different magnitude of bias updates. Solid curves are the median value, and dashed curves are the 33\% and 67\% percentiles of the errors. Both methods coincide for the rotation error as they use the same update rule. Solid and dashed curves coincide for the velocity and position errors. Our approach better approximates velocity and position preintegration updates. Statistics are computed over \num{e6} Monte-Carlo runs for each of the 166 preintegration factors. \label{fig:bias}}
\end{figure}

The derivation of the Jacobian is similar to the one we use to express the measurements as a large value plus a small perturbation. We first define the variation for one time step
\begin{align}
    \boUpsilon_{i}(\hbfb_i^+) &= \boUpsilon_{i}(\hbfb_i+\delta \bfb) \nonumber \\&=\hboUpsilon_{i}(\hbfb_i)\exp\big(\bfG_i \delta \bfb  + O(\|\delta\bfb\|^2)\big)
\end{align}
as similarly done in Section \ref{sec:justification}, where $\bfG_i$ is defined in \eqref{eq:G}. We then compute
\begin{align}
    \boUpsilon_{ij}(\hbfb_i^+) &\equal{eq:ups_batch}\prod_{k=i}^{j-1} \Phi_{k(j-1)}\left(\hboUpsilon_{k}(\hbfb_i) \exp\left(\bfG_k \delta \bfb\right) \right)\nonumber\\
    &\equal{eq:adjoint2}\hboUpsilon_{ij}(\hbfb_i) \prod_{k=i}^{j-1} \exp\left(\bfA_{k+1}^{j-1}\bfG_k \delta \bfb\right)\nonumber \\
    &\equal{eq:bch2}\hboUpsilon_{ij}(\hbfb_i)  \exp\Big(\sum_{k=i}^{j-1} \bfA_{k+1}^{j-1}\bfG_k \delta \bfb + O(\|\delta\bfb\|^2)\Big).
\end{align}
As in \eqref{eq:Sigma}, we resort  to \eqref{eq:sigmak1} to compute the Jacobian recursively as
\begin{align}
    \frac{\partial \hboUpsilon_{i(j+1)}}{\partial \bfb}|_{\hbfb_i} = \Ad_{\hboUpsilon_j}\bfF\frac{\partial \hboUpsilon_{ij}}{\partial \bfb}|_{\hbfb_i} + \bfG_j, \label{eq:Jbias}
\end{align}
starting from initial condition $\frac{\partial \hboUpsilon_{ii}}{\partial \bfb}|_{\hbfb_i} = \bfzero_{9\times6}$.

\subsection{Numerical Example with Biases}
\enbleu{This section shows our matrix formalism and the use of   exponential coordinates yield  slightly more accurate first-order bias correction than in the theory of \cite{forster_-manifold_2017} (with integration scheme of \cite{eckenhoff_closed-form_2019})}. The accuracy of this first-order bias correction is reported in Figure \ref{fig:bias}. To compute the statistics, we integrated the noise-free IMU measurements of Section \ref{sec:num_ex} with a given bias estimate $\hbfb_i=\bfzero_3$ during $\Delta t_{ij}=\SI{1}{s}$, which results in the preintegrated measurements $\Delta \bfR_{ij}(\hbfb_i)$, $\Delta \bfv_{ij}(\hbfb_i)$ and $\Delta \bfp_{ij}(\hbfb_i)$. Subsequently, a random perturbation $\delta \bfb$ with fixed magnitude was applied to both the gyroscope and accelerometer bias (the magnitude of accelerometer bias is 30 times higher than the gyro one). We repeat the integration at $\hbfb_i+\delta\bfb$ to obtain $\Delta \bfR_{ij}(\hbfb_i+\delta\bfb)$, $\Delta \bfv_{ij}(\hbfb_i+\delta\bfb)$ and $\Delta \bfp_{ij}(\hbfb_i+\delta\bfb)$. This ground-truth result was then compared against the first-order correction to compute the error of the approximation based on $N=\num{e6}$ random perturbation $\delta \bfb$ for each preintegration factor. The errors resulting from the first-order approximation are small, even for the relatively large bias perturbations, and we see our approach coincides with \cite{forster_-manifold_2017, eckenhoff_closed-form_2019} regarding orientation but obtains more accurate velocity and position preintegration factors.

\section{IMU Preintegration on Rotating Earth}\label{sec:coriolis}

This section provides exact closed-form expressions for computing the right-hand side of the factors \eqref{eq:drij}-\eqref{eq:prij} when taking into account the Earth rotation, Coriolis acceleration, and centrifugal acceleration. We obtain results that are independent of the uncertainty representation and as such may be used in previous preintegration formalisms \cite{forster_-manifold_2017,eckenhoff_closed-form_2019}. This allows   applying factor-graph based optimization techniques to military and civilian applications that require localization over long time scales based on accurate inertial sensors.

\subsection{IMU Equations with Rotating Earth}
Accounting for Earth rotation, \eqref{eq:dbfR}-\eqref{eq:dbfp} become \cite{farrell_aided_2008} 
\begin{align}
    \dbfR &= -\boOmega_\times\bfR + \bfR (\boomega^m - \bfb^{\boomega} - \boeta^{\boomega})_\times,  \label{eq:dbfR2} \\
    \dbfv &= \bfR(\bfa^m-\bfb^{\bfa}-\boeta^{\bfa}) + \bfg - 2\boOmega_\times \bfv - \boOmega_\times^2 \bfp, \\
    \dbfp &= \bfv, \label{eq:dbfp2}
\end{align}
where the Earth rotation vector
\begin{align}
    \boOmega = \underbrace{\text{Earth rate}}_{\simeq 7.292 \times 10^{-5} \text{rad/s}}\times\begin{bmatrix}\cos(\text{latitude})\\ 0\\-\sin(\text{latitude})\end{bmatrix} \in \bbR^3 \label{eq:rotvec}
\end{align}
is written in the local, i.e. geographic (north, east, down), reference frame, where the Earth rate is approximately \SI{15}{deg/h}. The term $-2\boOmega_\times \bfv$ is called Coriolis acceleration while the term $-\boOmega_\times^2 \bfp$ is called centrifugal acceleration. To be perfectly accurate, this second term is the varying part of the centrifugal acceleration, which actually writes $-\boOmega_\times^2 (\bfp-\bfp_0)$ with $\bfp_0$ a point of the Earth rotation axis. But expanding the parenthesis we obtain a constant term $\boOmega_\times^2 \bfp_0$ which can be simply added to $\bfg$. And this is already the case: the $\bfg$ of approximate value \SI{9.81}{m/s^2} we are familiar with is actually the sum of the Newton gravitation force and the centrifugal acceleration due to Earth rotation. Hence the residual term $-\boOmega_\times^2 \bfp$.

\enbleu{\begin{rmk}In the model \eqref{eq:dbfR2}-\eqref{eq:rotvec} and in the experiments of Section \ref{sec:experiments}, we consider  navigation in a local North-East-Down (NED) frame, where we assume the gravity $\bfg$ and Earth rate $\boOmega$ are fixed at their initial values. Although it simplifies the exposition, and is a valid assumption for trajectories that span a dozen of kilometers as in the following experiments, it is by no means a required assumption in our theory. Indeed, the only required assumption   is that $\bfg$, $\boOmega$ be fixed during the computation of each factor (which is   painless  as the latitude virtually does not change during one preintegration): at the beginning of each preintegration $\bfg$, $\boOmega$ may be updated based on the estimated position. 
In this regard, the algorithm needs to know at least its initial latitude. There is however no way around this assumption, which applies to all  algorithms which perform highly precise inertial navigation.\end{rmk}}

\begin{figure}
    \centering
    \includegraphics{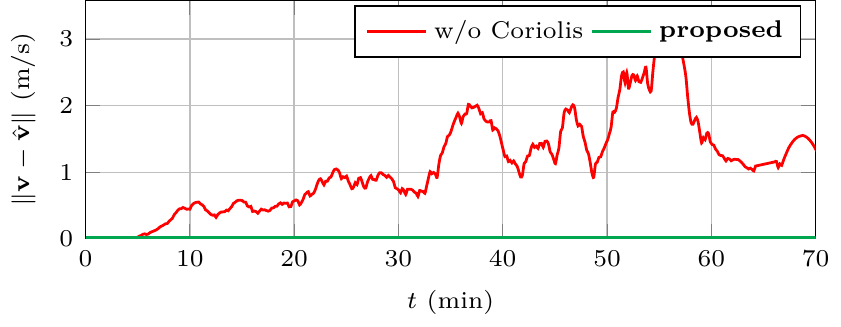}
    \caption{Error committed in term of velocity when neglecting rotating Earth and Coriolis accelerations (red), for a preintegration time interval $\Delta t_{ij} = \SI{5}{s}$. IMU measurements are bias- and noise-free, in simulation. The proposed preintegration approach (green) based on \eqref{eq:Rcoriolis}-\eqref{eq:pcoriolis} is exact. \label{fig:coriolis}}
\end{figure}

\subsection{Revisiting IMU Equations with Rotating Earth}
Model \eqref{eq:dbfR2}-\eqref{eq:dbfp2} does seemingly not lend itself to application of the form \eqref{eq:modelt}. However, let us introduce  the variable
\begin{align}
    \bfT' := \begin{bmatrix}\begin{array}{c|c|c}\bfR &\bfv + \boOmega_\times \bfp&\bfp\\\midrule\bfzero_{2\times3} & \multicolumn{2}{c}{\bfI_2} \end{array} \end{bmatrix}.
\end{align}
This trick allows embedding the auxiliary state $\bfT'_t$ into the form of \eqref{eq:modelt} as 
\begin{align}
    \bfT_{t}' = \boGamma_{t}' \Phi_{t}(\bfT_0') \boUpsilon_{t},
\end{align}
where $\boUpsilon_t$ is the same as in \eqref{eq:modelt} and $\boGamma'_t$ is obtained after solving the differential equations
\begin{align}
    \dboGamma_t' := \begin{bmatrix}\begin{array}{c|cc}\dboGamma^{\bfR} &\dboGamma^{\bfv}&\dboGamma^{\bfp}\\\midrule\bfzero_{2\times3} & \multicolumn{2}{c}{\bfI_2} \end{array} \end{bmatrix},
\end{align} 
where
\begin{align*}
    \dboGamma^{\bfR} = - \boOmega_\times \boGamma^{\bfR},~~
    \dboGamma^{\bfv} = \bfg - \boOmega_\times \boGamma^{\bfv},~~
    \dboGamma^{\bfp} = \boGamma^{\bfv}- \boOmega_\times \boGamma^{\bfp}, 
\end{align*}
initialized at $\boGamma^{\bfR} = \bfI_3$, $\boGamma^{\bfv}=\bfzero_3$,  $\boGamma^{\bfp}=\bfzero_3$, and that do not involve the state, see proofs in \cite{barrau_mathematical_2020} and  supplementary materials.  

We solve for $\bfT'_t$ and apply our framework for preintegration as follows. First, we define an exact discrete model
\begin{align}
    \bfT_{j}' = \boGamma_{ij}' \Phi_{ij}(\bfT_i') \boUpsilon_{ij}, \label{eq:tij}
\end{align}
between instants $t_i$ and $t_j$, as performed in Section \ref{sec:batch}, that leads to
\begin{align}
    \bfR_{j} &= \boGamma^{\bfR}_{ij} \bfR_i \Delta \bfR_{ij}, \label{eq:Rcoriolis} \\
    \bfv_j &= \boGamma^{\bfv}_{ij} + \boGamma^{\bfR}_{ij}\left(\bfR_i\Delta \bfv_{ij} + \bfv_i + \boOmega_{\times}\bfp_i\right) - \boOmega_{\times}\bfp_j, \label{eq:vcoriolis}\\
    \bfp_j &= \boGamma^{\bfp}_{ij} + \boGamma^{\bfR}_{ij}\left(\bfR_i\Delta \bfp_{ij} + \left(\bfv_i+\boOmega_{\times}\bfp_i\right)\Delta t_{ij} + \bfp_i\right),\label{eq:pcoriolis}
\end{align}
where $\Delta \bfR_{ij}$, $\Delta \bfv_{ij}$, $\Delta \bfp_{ij}$ are yet given through \eqref{eq:DeltaR}-\eqref{eq:Deltap} while $\boGamma^{\bfR}_{ij}$, $\boGamma^{\bfv}_{ij}$, $\boGamma^{\bfp}_{ij}$ are defined as
\begin{align}
    \boGamma^{\bfR}_{ij} &= \exp(-\Delta t_{ij} \boOmega), \\
    \boGamma^{\bfv}_{ij} &= \calJ_{-\Delta t_{ij} \boOmega} \Delta t_{ij}\bfg, \\
    \boGamma^{\bfp}_{ij} &=  \big(\frac{\Delta t^2}{2}\bfI_3 + a\boOmega_\times + b\boOmega_\times^2\big)\bfg,
\end{align}
with $\phi=\|\boOmega\|$, $a=\phi^{-3}(\phi \Delta t_{ij}\cos(\phi \Delta t_{ij}) - \sin(\phi \Delta t_{ij}))$, and $b=\phi^{-4}(\frac{\phi^2\Delta t_{ij}^2}{2} - \cos(\phi \Delta t_{ij}) - \phi \Delta t_{ij} \sin(\Delta t_{ij}\phi) + 1)$. To express preintegration factors as function of states, we compute
\begin{align}
    \boUpsilon_{ij} &\equal{eq:tij}  \Big(\boGamma'_{ij} \Phi_{ij}\left(\bfT'_i\right)\Big)^{-1}\bfT'_{j}, \label{eq:preint_mes_cor} 
\end{align}
see the similarity with \eqref{eq:preint_mes}, that we finally develop as
\begin{align}
    \Delta \bfR_{ij} &\equal{eq:Rcoriolis} \big(\boGamma^{\bfR}_{ij} \bfR_i\big)^T \bfR_{j} ,\label{eq:deltaR2}\\
    \Delta \bfv_{ij} &\equal{eq:vcoriolis}  \bfR_i^T\big(\boGamma^{\bfR T}_{ij}\left(\bfv_j+\boOmega_{\times}\bfp_j-\boGamma_{ij}^{\bfv}\right)-\bfv_i -\boOmega_{\times}\bfp_i \big),\\
    \Delta \bfp_{ij} &\equal{eq:pcoriolis} \bfR_i^T\big(\boGamma^{\bfR T}_{ij}\left(\bfp_j-\boGamma^{\bfp}_{ij}\right) - \left(\bfv_i+\boOmega_{\times}\bfp_i\right)\Delta t_{ij} - \bfp_i\big). \label{eq:deltap2}
\end{align}
To summarize, preintegrating IMU with rotating Earth consists in substituting \eqref{eq:drij}-\eqref{eq:prij} by \eqref{eq:deltaR2}-\eqref{eq:deltap2}, and leaving the covariance noise computation unchanged, as Coriolis and centrifugal accelerations leave the left-hand side of the preintegrated factors unchanged, and thus the quantity $\boUpsilon_{ij}$ in \eqref{eq:preint_mes_cor} is yet defined in  \eqref{eq:ups_pre}. \enbleu{These results are obtained independently from a choice for uncertainty representation, and as such they may also be used in current implementations such as \cite{forster_-manifold_2017}.}

We confirm our results with a numerical example, see Figure \ref{fig:coriolis}, where we preintegrate bias- and noise-free IMU measurements for a long car trajectory. Similarly as in Section \ref{sec:num_ex}, we differentiate the ground-truth trajectory of the sequence 1 of our real experiments (see Section \ref{sec:experiments}) to obtain true inertial measurements. Then the ground-truth is recomputed with model \eqref{eq:dbfR2}-\eqref{eq:dbfp2} to remove integration errors. We compute the error when neglecting the rotation of the Earth ($\boOmega=\bfzero_3$) versus our approach with $\Delta t_{ij} = \SI{5}{s}$. We see no error at the beginning of the sequence as the car is static. Then the error committed when neglecting the rotating Earth grows, whereas our approach is \emph{exact}.

\begin{rmk} \cite{indelman_factor_2012, indelman_information_2013} attack factor-graph based accurate navigation, and provide the following formulas for preintegration with Coriolis acceleration in the appendix of \cite{indelman_factor_2012}\label{rmk:coriolis}
\begin{align*}
    v_j^{L_j}= R^{L^j}_{L_i}\Big(v_i^{L_i}+ R^{L_i}_{b_i}\Delta v^{b_i}_{i \rightarrow j} + [g^{L_i}-\frac{(w_{iL_i}^{L_i})_\times v_j^{L_j}}{2}]\Delta t_{ij} \Big).
\end{align*}
The authors obtain a term $-2 (\omega_{i^L_i})_\times v_i^{L_i}$ in place of the expected $-2 (\omega_{i^L_i})_\times v_j^{L_j}$ (index $i$ of $v$ should be $j$): we see the Coriolis term of \cite{indelman_factor_2012} is actually approximated by its value at initial time $t_i$.
\end{rmk}

\renewcommand{\figurename}{Table}
\setcounter{figure}{0} 
\begin{figure}
    \centering
    \footnotesize
    \begin{tabular}{c||c|c}
		\toprule
	    IMU noise parameters & \textbf{ours} & simu. of \cite{forster_-manifold_2017} \\\midrule
        gyro rand. walk, \si{$\text{rad/(s}\sqrt{\text{Hz}}\text{)}$} & \num{9.4e-6} & \num{7e-4}\\\midrule 
        acc. rand. walk, \si{$\text{m/(s}^2\sqrt{\text{Hz}}\text{)}$}& \num{4.2e-3} & \num{1.9e-2}\\\midrule 
        gyro bias rand. walk, \si{$\text{rad/(s}^2\sqrt{\text{Hz}}\text{)}$}& \num{1.6e-6} & \num{4e-4}\\\midrule 
        acc. bias rand. walk, \si{$\text{m/(s}^3\sqrt{\text{Hz}}\text{)}$}&\num{5.4e-5} & \num{1.2e-2} 
        \\
		\bottomrule
	\end{tabular}
    \caption{Values of the IMU noise parameters used in our algorithms during real experiments, compared to the ones of \cite{forster_-manifold_2017}.  \label{fig:imu_spec}}
\end{figure}
\renewcommand{\figurename}{Fig.}
\setcounter{figure}{10} 

\renewcommand{\figurename}{Table}
\setcounter{figure}{1} 
\begin{figure*}
    \centering
    \footnotesize
    \begin{tabular}{c||c|c||c|c||c|c||c|c}
		\toprule
		 seq.  & length & dur. &  \multicolumn{2}{c||}{LiDAR}& \multicolumn{2}{c||}{LiDAR-IMU w/o Coriolis} &\multicolumn{2}{c}{LiDAR-IMU w Coriolis (\textbf{proposed})}\\\midrule
		&  km & min  &  rot. err. (\si{deg/km}) &  trans. err. (\si{m/km}) &  rot. err. (\si{deg/km}) &  trans. err. (\si{m/km}) & rot. err. (\si{deg/km}) &  trans. err. (\si{m/km}) \\\midrule
		1 &27 &73 &1.6~~~/~~~0.9&15~~~/~~~19&1.3~~~/~~~0.6&\textbf{15}~~~/~~~\ 8&\textbf{0.9}~~~/~~~\textbf{0.3}&\textbf{15}~~~/~~~\ \textbf{6}\\
        2 &37 &101&1.5~~~/~~~0.9&14~~~/~~~12&\textbf{0.9}~~~/~~~0.6&\textbf{13}~~~/~~~10&0.6~~~/~~~\textbf{0.3}&\textbf{13}~~~/~~~\ \textbf{7}\\
        3 &3 &7 &3.0~~~/~~~1.7&16~~~/~~~\ 9&1.5~~~/~~\textbf{~0.2}&\textbf{14}~~~/~~~\ \textbf{4}&\textbf{1.4}~~~/~~~0.3&15~~~/~~~\ \textbf{4}\\
        4 &8 &19 &1.7~~~/~~~1.1&19~~~/~~~22&\textbf{0.8}~~~/~~~\textbf{0.3}&18~~~/~~~11&1.1~~~/~~~\textbf{0.3}&\textbf{17}~~~/~~~\ \textbf{6}\\
        5 &4 &11 &2.7~~~/~~~1.6&16~~~/~~~\ 9&\textbf{1.0}~~~/~~~\textbf{0.4}&\textbf{16}~~~/~~~\ \textbf{5}&\textbf{1.0}~~~/~~~0.5&\textbf{16}~~~/~~~\ \textbf{5}\\
        6 &74 &73 &3.9~~~/~~~0.9&37~~~/~~~26&\textbf{2.6}~~~/~~~\textbf{0.4}&38~~~/~~~25&\textbf{2.6}~~~/~~~\textbf{0.4}&\textbf{35}~~~/~~~\textbf{24}\\
		\midrule[1pt]
		 total & 153 & 284& 2.6~~~/~~~0.9&25~~~/~~~19&1.7~~~/~~~0.5&25~~~/~~~16&\textbf{1.6}~~~/~~~\textbf{0.3}&\textbf{23}~~~/~~~\textbf{14}\\
		\bottomrule
	\end{tabular}
    \caption{Real experiment results in terms of relative translation (trans. err.) and relative rotational (rot. err.) errors based on short (left) and long (right) sub-sequences. Taking into account IMU during estimation improves each metric. Considering rotating Earth and Coriolis acceleration is more beneficial for long-term navigation.    \label{fig:exp_res}}
\end{figure*}
\renewcommand{\figurename}{Fig.}
\setcounter{figure}{10} 
\section{Real Experiments}\label{sec:experiments}

This section addresses  sensor-fusion of a high-grade IMU with relative translations provided by a LiDAR in a fixed-lag smoother for long-term navigation. Building on the preintegration with rotating Earth and Coriolis effect formulas developed in Section \ref{sec:coriolis}, we implement our loosely-coupled approach with the GTSAM factor-graph library \cite{dellaert_factor_2012}.

We gathered more than 150 km and four hours of data divided into six sequences. The vehicle acquires data around the R\&T centre of SafranTech\footnote{SafranTech is a research centre of the company Safran, a French multinational aircraft engine, rocket engine, aerospace-component and defense company. The second author is working there as an engineer.} located at Magny-Les-Hameaux, France. The car embeds a high-grade IMU manufactured by   Safran Electronics and Defense, whose increments are acquired at \SI{100}{Hz}, and a Velodyne LiDAR VLP32C which is mounted on top of the car. Table \ref{fig:imu_spec} indicates the IMU noise specifications that we set in our algorithms and which are two orders of magnitude lower than those used in the simulations of \cite{forster_-manifold_2017}. The 3D laser scans between keyframes are preprocessed to obtain relative transformations using scan matching algorithms. The vehicle is finally equipped with a RTK-GPS antenna which outputs positions and yaws that we consider as ground-truth. This dataset is challenging regarding the length of  the sequences, the car velocity (up to \SI{30}{m/s}), and the presence of many bumps, sharp curves and roundabouts on the way.

\subsection{Compared Methods \& Algorithm Setting}
Owing the precision of the gyroscopes, relative orientation between LiDAR's point clouds does not bring additional relevant information, and we focus only on the relative translations  returned by the LiDAR.  We compare three approaches for estimating the vehicle pose using the IMU and the LiDAR:
\begin{enumerate}
    \item \textbf{LiDAR}, which is the compound of the relative poses estimated by the scan matching algorithms as in \eqref{eq:com_pos};
    \item \textbf{LiDAR-IMU w/o Coriolis}, which is the fusion between the relative translations given by the scan matching algorithms and the inertial signals, without considering rotating Earth and Coriolis force, i.e. assuming $\boOmega=\bfzero_3$;
    \item \textbf{LiDAR-IMU w Coriolis}, same as method 2), but where rotating Earth and Coriolis force are considered and $\boOmega$ defined with the latitude at which each sequence starts (\SI{48.73}{deg}).
\end{enumerate}
The LiDAR-aided approaches are described as follows. We implement a fixed-lag smoother based on iSAM2 \cite{kaess_isam2_2012} from the GTSAM library \cite{dellaert_factor_2012}. Our GTSAM fork\footnote{Our GTSAM repo is available at \texttt{\url{https://github.com/mbrossar/gtsam}}. We also modify the initial Coriolis effect correction, see Remark \ref{rmk:coriolis},  which proved to be bugged.} computes the bias update correction with Lie exponential coordinates \eqref{eq:biasupdate}, preintegrates with rotating Earth and Coriolis force \eqref{eq:Rcoriolis}-\eqref{eq:pcoriolis}, and leaves the retraction of the extended pose (a.k.a. \texttt{navigation state} in \cite{dellaert_factor_2012}) unchanged. Biases are estimated along with extended poses through \texttt{CombinedImuFactor}, and initialized with the beginning of each sequence, where the car is static. Each approach performs odometry estimation, i.e., it estimates the car trajectory relative to its starting position without using global information from GNSS or LiDAR loop-closure.

Scan matching algorithms return relative orientation and translation, where the level of rotation uncertainty between LiDAR scans is much higher than the gyro’s uncertainty (assuming the gyro bias is known, the gyro only drifts of less than a few degrees in one hour, which is lower than the Earth rotation rate). Therefore, we fed the factor-graph with IMU preintegrated factors, relative translation factors, and zero upward velocity factor with a relatively small standard deviation of \SI{1}{m/s} to prevent the upward velocity from drifting, as used, e.g., in \cite{ye_tightly_2019}.

We set the smoother with a lag of \SI{20}{s}, insert each factor at \SI{4}{Hz}, and define its noise parameters as follows. Then the covariance of the discrete-time noise $\boeta^{\mathrm{d}}$ is computed as a function of the sampling rate and relates to the continuous-time spectral noise $\boeta^{\mathrm{c}}$ via $\cov(\boeta^{\mathrm{d}}) = \cov(\boeta^{\mathrm{c}})/\Delta t$ \cite{crassidis_sigma_2006}. Relative translations are given with an uncertainty of \SI{20}{cm} in each direction, and we make the translation factor more robust during high motions with a Huber loss. Finally, the initial orientation and position are provided from the ground-truth, and the initial velocity is null.

\subsection{Evaluation Metrics}
To assess performances we recall the error metrics proposed in the KITTI dataset \cite{geiger_vision_2013}:
\begin{itemize}
    \item \textbf{Relative translation error}, which is computed as 
    \begin{align*}
       \text{trans. err.} = \left\|\bfR_i^T(\bfp_j-\bfp_i) - \hbfR_i^T(\hbfp_j-\hbfp_i)\right\|_2
    \end{align*}
    after an alignment transformation for all sub-sequences of length \SI{100}{m}, \ldots, \SI{800}{m};
    \item \textbf{Relative rotational error}, which is computed as 
    \begin{align*}
        \text{rot. err.} = \left\|\log\big( (\bfR_i^T\bfR_j)^T (\hbfR_i^T\hbfR_j) \big) \right\|_2
     \end{align*}
     for all sub-sequences of length \SI{100}{m}, \ldots, \SI{800}{m}.
\end{itemize}

As advocated in \cite{zhang_tutorial_2018}, these metrics are recommended for comparing  odometry estimation methods since they are barely sensitive to the time when  the estimation error occurs. We adapt theses metrics as follows: the approaches are evaluated in the horizontal plan as the ground-truth is as accurate as the estimation algorithms to compute the pitch and the roll, and we compute both short- and long-term metrics. In total, we compute three sets of pairs \{rot. err., trans. err.\}:   
\begin{enumerate}
    \item a set of short-term errors based on sub-sequences of length \SI{100}{m}, \ldots, \SI{800}{m} as in \cite{geiger_vision_2013};
    \item a set of long-term errors based on sub-sequences of length \SI{1}{km}, \ldots, \SI{8}{km}, as rotating Earth and Coriolis effect are more visible after long traveled distances;
    \item a set of errors for sub-sequences of duration \SI{3}{min}, \ldots, \SI{15}{min}. Indeed, the error of odometry methods based on LiDAR or vision should grow with distance whereas that methods based on IMU should grow with time.
\end{enumerate}

\subsection{Result Analysis \& Discussion}
\begin{figure}
    \centering
    \includegraphics{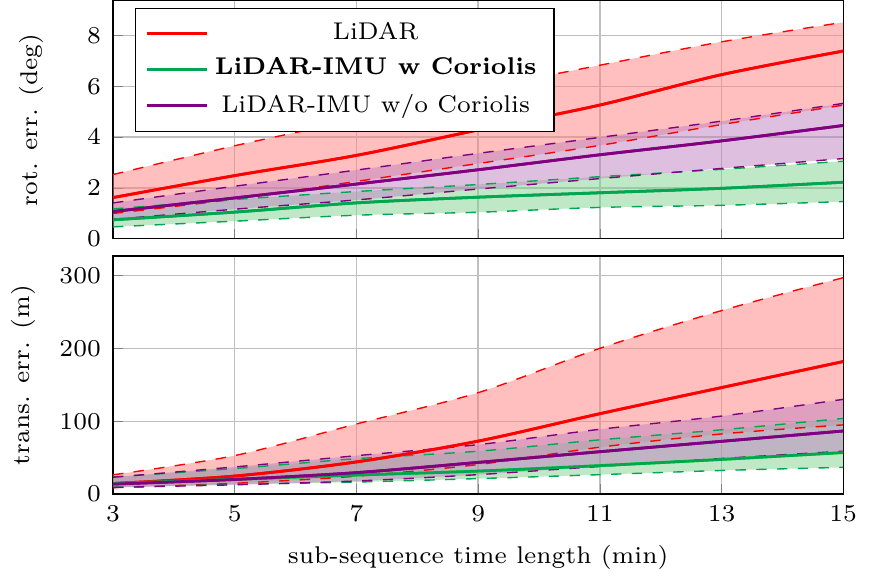}
    \caption{Real experiment results in terms of relative rotation (rot. err.) and relative translation (trans. err.) errors based on sub-sequences of duration 3, \ldots, 15 minutes. Solid curves are the median value, and dashed curves are the 33\% and 67\% percentiles of the errors. Considering rotating Earth and Coriolis force appears beneficial for orientation and translation from sub-sequences of length 7 minutes.\label{fig:exp_res2}}
\end{figure}

Table \ref{fig:exp_res} provides numerical results for the two first sets of evaluation metrics, and Figure \ref{fig:exp_res2} displays results for the third set. We observe:
\begin{enumerate}
    \item the LiDAR estimates highly efficiently the distances and stays accurate in straight lines. However, it has difficulties in areas with tight curves, roundabouts (where visual ambiguity might play a role), or when the car is moving fast;
    \item as the LiDAR loses accuracy in decisive but rare moments such as roundabouts, the short-term results of Table \ref{fig:exp_res} are correct, and one mainly tries to improve the long-term metrics;
    \item only relative translation information is sufficient to obtain a robust loosely-coupled estimation from the IMU and the LiDAR. This solves most of the problems encountered by the LiDAR during sharp curves and roundabouts;
    \item  the two LiDAR inertial approaches have similar results for the shortest sequences 3 and 5, see Table \ref{fig:exp_res}. Indeed, for a small sequence, Coriolis force is negligible and the Earth rotation rate may be inserted in bias;
    \item taking into account Coriolis increases accuracy for long distances, see the long-term metrics in Table \ref{fig:exp_res}, where the one that considers Coriolis is about one hundred meter away from the ground after more than one hour of odometry, without GNSS or LiDAR loop-closure;
    \item Figure \ref{fig:exp_res2} shows approximately linear grow  of the error with respect to time. Considering rotating Earth and Coriolis effect improves the translation metric from \SI{130}{m} to \SI{103}{m}, i.e. an improvement of 21\%, with sub-sequences of length \SI{15}{min}.
\end{enumerate}

These results show how beneficial it is to consider rotating Earth and Coriolis force in IMU preintegration theory. However, our experiments do not show the advantage of the Lie exponential update of the bias. Indeed, the differences between the proposed bias update and the standard one remains below, e.g., parameter tuning. The benefit  of bias update with exponential Lie update does not necessary require long sequences but more accurate localization systems and larger preintegration times.

In our experiments, we take into account the rotation of the Earth but not its curvature (otherwise it would be necessary to consider the variations of longitude, latitude, ellipsoidal altitude, and transport rate). This point is justified by the relatively small area covered by the trajectories, and the fact that we are estimating relative trajectories. We anticipate the proposed approach would provide even more improvements for long-term navigation with absolute measurements and long distances, e.g. for a plane, a drone, or   autonomous underwater vehicles equipped with accurate inertial sensors.

\section{Conclusion}\label{sec:conclu}

This paper presents some generic techniques to associate uncertainty to an extended pose (three dimensional orientation, velocity and position of a rigid body) and shows how to propagate associated uncertainties with fourth-order accuracy in noise variables and second order in the associated covariances. The framework additionally provides an elegant mathematical approach that brings further maturity to the theory of preintegration on manifold. It unifies flat and rotating Earth  IMU equations within a single framework, hence providing  extensions of the theory of preintegration with Coriolis force. The method compares favorably against state-of-the-art in extensive simulations, and has been validated for one hour long car navigation in an efficient fixed-lag smoother that fuses IMU and relative translations provided by a LiDAR. 

Looking forward, we believe these techniques open up for novel implementations of factor-graph based methods to the context of long term inertial-aided navigation systems, hence genuine industrial navigation systems.  Finally, the theory could find application in other problems requiring detailed bookkeeping of extended pose uncertainties. Indeed, our approach with fine uncertainty representation may prove  decisive for finely detecting GNSS outliers with preintegration\cite{Chng_outlier-robust_2019}, or when we search to accurately define the uncertainty of distant relative extended poses  in an (extended) pose-graph. \enbleu{This is outlined by the example of Section \ref{sec:simple_exm} where our approach improves the mean and the uncertainty of the estimates of existing approaches it is compared to.}
\bibliographystyle{IEEEtran}
\bibliography{biblio}
\appendix
\subsection{$SO(3)$ \& $SE_2(3)$ Closed-Form Expressions}
Let $\bophi \in \bbR^3$ and $\bfR \in SO(3)$. The exponential, logarithm, left-Jacobian and inverse left-Jacobian of $SO(3)$ are given as
\begin{align}
    \exp(\bophi) &= \bfI_3 + \frac{\sin \phi}{\phi}\bophi_\times + \frac{1-\cos \phi}{ \phi^2}\bophi_\times^2,\\
    \log(\bfR) &= \frac{\varphi }{2\sin \varphi}(\bfR-\bfR^T)^\vee,\\ 
    \calJ_{\bophi} &=\bfI_3 + \frac{1-\cos \phi}{ \phi^2}\bophi_\times + \frac{\phi-\sin \phi}{ \phi^3}\bophi_\times^2, \\
    \calJ^{-1}_{\bophi} &= \bfI_3 - \frac{1}{2}\bophi_\times + \left( \phi^{-2}+\frac{1+\cos \phi}{2\phi\sin \phi}\right)\bophi_\times^2,
\end{align}
where $\vee$ is the linear inverse operator of $\wedge$, $\phi=\|\bophi\|$ and $\varphi=\cos^{-1}(\frac{\trace(\bfR)-1}{2})$. Let $\boxi \in \bbR^9$ and $\bfT\in SE_2(3)$. The exponential and logarithm of $SE_2(3)$ are computed in overloaded operators as
\begin{equation}
    \exp(\boxi) = \begin{bmatrix}
        \exp(\bophi) & \calJ_{\bophi} \bonu & \calJ_{\bophi} \borho \\
        \bfzero_3^T & 1& 0 \\
        \bfzero_3^T & 0& 1
    \end{bmatrix}, \hfill
    \log(\bfT) = \begin{bmatrix}\log(\bfR)\\
        \calJ^{-1}_{\log(\bfR)} \bonu \\
        \calJ^{-1}_{\log(\bfR)} \borho
    \end{bmatrix}.
\end{equation}
The Jacobian for $SE_2(3)$ and its inverse are derived from those for $SE(3)$ introduced in \cite{barfoot_associating_2014} as
\begin{align}
    \calJ_{\boxi} &:= \begin{bmatrix}
        \calJ_{\bophi} & \bfzero_{3\times3}& \bfzero_{3\times3}\\
        \calQ_{\bophi,\bonu} & \calJ_{\bophi}& \bfzero_{3\times3}\\
        \calQ_{\bophi,\borho}& \bfzero_{3\times3} & \calJ_{\bophi}
        \end{bmatrix},\\
    \calJ^{-1}_{\boxi} &:= \begin{bmatrix}
        \calJ^{-1}_{\bophi} & \bfzero_{3\times3}& \bfzero_{3\times3}\\
        -\calJ^{-1}_{\bophi}\calQ_{\bophi,\bonu}\calJ^{-1}_{\bophi}& \calJ^{-1}_{\bophi}& \bfzero_{3\times3}\\
        -\calJ^{-1}_{\bophi}\calQ_{\bophi,\borho}\calJ^{-1}_{\bophi}& \bfzero_{3\times3} & \calJ^{-1}_{\bophi}
        \end{bmatrix},
\end{align}
\begin{align}
    \calQ&_{\bophi,\bonu}: = \frac{1}{2}\bonu_\times + \frac{\phi - \sin\phi}{\phi^3}\left(\bophi_\times \bonu_\times + \bonu_\times \bophi_\times + \bophi_\times \bonu_\times \bophi_\times\right)\nonumber \\
    &+\frac{\phi^2+2\cos\phi-2}{\phi^4}\left(\bophi_\times \bophi_\times \bonu_\times + \bonu_\times \bophi_\times \bophi_\times - 3 \bophi_\times \bonu_\times \bophi_\times\right) \nonumber\\
    &+\frac{2\phi-3\sin\phi+\phi\cos\phi}{2\phi^5} \left(\bophi_\times \bonu_\times \bophi_\times \bophi_\times + \bophi_\times \bophi_\times \bonu_\times \bophi_\times\right),\label{eq:calQ}
\end{align}
and $\calQ_{\bophi,\borho}$ is defined similarly as \eqref{eq:calQ}, replacing $\bonu$ by $\borho$.

\subsection{Third- and Fourth-Order Contributions}
We compute the quantity $\bfS_{\text{4th}}$ in \eqref{eq:sigmak1} along the lines of \cite{barfoot_associating_2014}. Les us define the operators
\begin{align}
    \ll\bfA\gg &:= -\trace(\bfA)\bfI_3 + \bfA, \\
    \ll\bfA,\bfB\gg &:= \ll\bfA\gg\ll\bfB\gg + \ll\bfB,\bfA\gg,
\end{align}
and $\boSigma = \bfA_i \boSigma_{i+1}\bfA_i^T$. We obtain
\begin{align}
    \bfS_{\text{4th}} = \frac{1}{12}\left(\bfA_{\boSigma} \bfQ + \bfQ \bfA_{\boSigma}^T + \bfA_{\bfQ} \boSigma + \boSigma \bfA_{\bfQ}^T\right) + \frac{1}{4}\bfB,
\end{align}
where
\begin{align*}
    \bfA_{\boSigma} &= \begin{bmatrix}
        \ll\boSigma_{\bophi \bophi}\gg &\bfzero_{3\times3}&\bfzero_{3\times3}\\
        \ll\boSigma_{\bonu \bophi} + \boSigma_{\bophi \bonu}\gg & \ll\boSigma_{\bophi \bophi}\gg &\bfzero_{3\times3} \\
        \ll\boSigma_{\borho \bophi} + \boSigma_{\bophi \borho}\gg&\bfzero_{3\times3}&\ll\boSigma_{\bophi \bophi}\gg,
    \end{bmatrix}, \\
    \bfA_{\bfQ} &= \begin{bmatrix}
        \ll\bfQ_{\bophi \bophi}\gg &\bfzero_{3\times3}&\bfzero_{3\times3}\\
        \ll\bfQ_{\bonu \bophi} + \bfQ_{\bophi \bonu}\gg & \ll\bfQ_{\bophi \bophi}\gg &\bfzero_{3\times3} \\
        \ll\bfQ_{\borho \bophi} + \bfQ_{\bophi\borho}\gg&\bfzero_{3\times3}&\ll\bfQ_{\bophi \bophi}\gg\end{bmatrix},
    \end{align*}
    \begin{align}        
    \bfB_{\bophi\bophi}&= \ll\boSigma_{\bophi \bophi}, \bfQ_{\bophi \bophi}\gg,\\
    \bfB_{\bonu\bophi}&= \bfB_{\bophi\bonu}^T = \ll\boSigma_{\bophi \bophi}, \bfQ_{\bophi\bonu}\gg + \ll\boSigma_{\bonu\bophi}, \bfQ_{\bophi \bophi}\gg,\\
    \bfB_{\borho\bophi}&= \bfB_{\bophi\borho}^T = \ll\boSigma_{\bophi \bophi}, \bfQ_{\bophi\borho}\gg + \ll\boSigma_{\borho\bophi}, \bfQ_{\bophi \bophi}\gg,\\
    \bfB_{\bonu\bonu}&= \ll\boSigma_{\bophi \bophi}, \bfQ_{\bonu \bonu}\gg + \ll\boSigma_{\bophi\bonu}, \bfQ_{\bonu \bophi}\gg \nonumber \\
    &+ \ll\boSigma_{\bonu \bophi}, \bfQ_{\bonu \bophi}\gg + \ll\boSigma_{\bonu \bonu}, \bfQ_{\bophi \bophi}\gg,\\
    \bfB_{\bonu\borho}&= \bfB_{\borho\bonu}^T = \ll\boSigma_{\bonu \borho}, \bfQ_{\bophi \bophi}\gg + \ll\boSigma_{\bophi \borho}, \bfQ_{\bonu \bophi}\gg \nonumber \\
    &+ \ll\boSigma_{\bonu\bophi}, \bfQ_{\borho\bophi}\gg + \ll\boSigma_{\bophi \bophi}, \bfQ_{\bonu \borho}\gg,\\
    \bfB_{\borho\borho}&=\ll\boSigma_{\bophi \bophi}, \bfQ_{\borho \borho}\gg + \ll\boSigma_{\bophi\borho}, \bfQ_{\borho \bophi}\gg \nonumber \\
    &+ \ll\boSigma_{\borho \bophi}, \bfQ_{\bophi\borho}\gg + \ll\boSigma_{\borho \borho}, \bfQ_{\bophi \bophi}\gg.
\end{align}

\end{document}